\documentclass[pdflatex,sn-mathphys-num]{sn-jnl}
\renewcommand{\arraystretch}{1.05}
\usepackage{xcolor}
\usepackage{array}
\usepackage{subcaption}
\renewcommand{\arraystretch}{1.05}    
\usepackage{times}
\usepackage{latexsym}
\usepackage{graphicx}
\usepackage{siunitx}       
\usepackage{threeparttable}
\usepackage{tcolorbox}
\tcbuselibrary{breakable}
\usepackage[noend]{algpseudocode}
\usepackage[T1]{fontenc}

\usepackage[utf8]{inputenc}

\usepackage{microtype}
\usepackage{inconsolata}

\usepackage{booktabs}
\usepackage{multirow}%
\usepackage{amsmath,amssymb,amsfonts}%
\usepackage{amsthm}%
\usepackage{mathrsfs}%
\usepackage[title]{appendix}%
\usepackage{textcomp}%
\usepackage{manyfoot}%
\usepackage{algorithm}%
\usepackage{algorithmicx}%
\usepackage{listings}%

\theoremstyle{thmstyleone}%
\theoremstyle{thmstyletwo}%

\theoremstyle{thmstylethree}%

\begin{document}

\title[\textsc{PC-MoE}: Memory-Efficient and Privacy-Preserving Collaborative Training for Mixture-of-Experts LLMs]{\textsc{PC-MoE}: Memory-Efficient and Privacy-Preserving Collaborative Training for Mixture-of-Experts LLMs}


\author*[1,2]{\fnm{Ze Yu} \sur{ Zhang}}\email{zhan1130@comp.nus.edu.sg}

\author[2]{\fnm{Bolin}
\sur{Ding}}\email{bolin.ding@alibaba-inc.com}

\author[1]{\fnm{Bryan Kian Hsiang} \sur{Low}}\email{lowkh@comp.nus.edu.sg}

\affil[1]{\orgdiv{Department of Computer Science}, \orgname{National University of Singapore}, \city{Singapore}, \country{Republic of Singapore}}

\affil[2]{\orgname{Alibaba Group}, \city{Singapore}, \country{Republic of Singapore}}

\abstract{Mixture-of-Experts (MoE) has been gaining popularity due to its successful adaptation to large language models (LLMs). In this work, we introduce \emph{Privacy-preserving Collaborative Mixture-of-Experts (PC-MoE)}, which leverages the sparsity of the MoE architecture for \emph{memory-efficient} decentralized collaborative LLM training, enabling multiple parties with limited GPU-memory and data resources to collectively train more capable LLMs than they could achieve individually. At the same time, this approach protects training data privacy of each participant by keeping training data, as well as parts of the forward pass signal and gradients locally within each party. By design, \textsc{PC-MoE} synergistically combines the strengths of distributed computation with strong confidentiality assurances. Unlike most privacy-preserving schemes, which pay for confidentiality with lower task accuracy, our framework \emph{breaks that trade-off}: across seven popular LLM benchmarks, it almost matches (and sometimes exceeds) the performance and convergence rate of a fully centralized model, enjoys near 70\% peak GPU RAM reduction, while being fully robust against reconstruction attacks.}




\maketitle

\section{Introduction}

Large language models (LLMs) owe many of their most striking abilities—few-shot reasoning, multilingual competence, tool use—to sheer scale: once parameter counts and training-data volumes pass certain thresholds, qualitatively new behaviors emerge \cite{brown2020language,wei2022emergent,kojima2022large,kaplan2020scaling,hoffmann2022training}.  However, reaching those thresholds first requires hardware that most teams simply cannot touch: multi-TB GPU memory, petabyte-per-day interconnects, and clusters large enough to idle entire racks for weeks \cite{thompson2020computational,patterson2021carbon}. This hardware gate places frontier-level experimentation well beyond the reach of most academic labs, start-ups, and public-interest organizations, potentially concentrating progress in the hands of a few well-funded actors. How can the community reap the benefits of scale without replicating its prohibitive hardware and data footprint?

Mixture-of-Experts (MoE) LLMs address the compute bottleneck by routing each token through only a handful of specialist sub-networks, keeping the \emph{per-token} FLOPs nearly constant even as the total parameter count grows \cite{shazeer2017outrageously,lepikhin2020gshard,fedus2022switch,du2022glam}.  Beyond raw arithmetic savings, this sparsity localizes most matrix multiplications, which in turn eases the three dominant hardware constraints—on-card memory, GPU-to-GPU bandwidth, and cluster interconnect contention.  Recent systems such as DeepSeek-V3 demonstrate that, when combined with ultra-low-precision training (FP8) and topology-aware networking, MoE enables trillion-parameter LLMs without blowing up the per-device budget, making it the de facto architecture for practical large-scale scaling \cite{zhao2025insights,rajbhandari2022deepspeed}.

Yet the total parameter count in an MoE LLM still remains enormous—expert weights routinely account for the majority of all trainable parameters. Consequently, training even a \emph{sparse} LLM still demands high-memory GPUs, and it \emph{still} assumes access to diverse, high-quality data \cite{shoeybi2019megatron}. Many potential contributors possess only a modest amount of either resource. A natural remedy is to let multiple parties pool their limited compute \emph{and} complementary datasets to co-train a single model that none could manage alone \cite{mcmahan2017communication, borzunov2022petals}. But a naïve collaborative scheme re-introduces the very drawbacks it is meant to avoid: each participant may have to juggle massive weight tensors, and sensitive training data can leak through shared gradients or parameters \cite{zhu2019deep,geiping2020inverting}.

\begin{figure*}[!ht]
    \centering
    \includegraphics[width=0.8\linewidth]{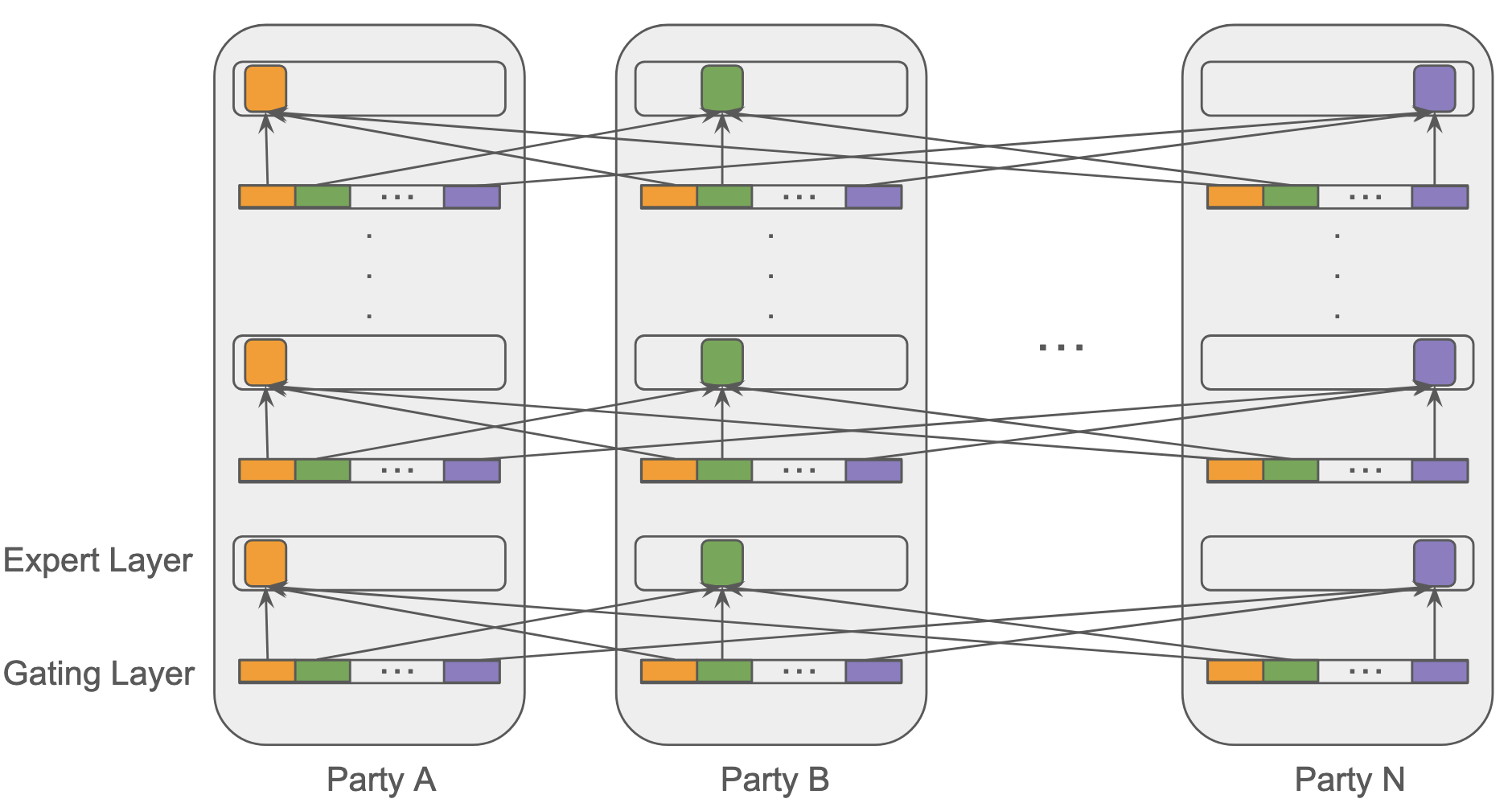}
    \caption{
         Overall architecture of the distributed MoE training framework. Each colored block in the expert layer represents the subset of experts in that layer belonging to the party, whereas the block in the gating layer with the same color represents its corresponding activation. During forward pass, the top-k activated experts do not necessarily belong to the same colored block, therefore some of the signal will be passed to the experts of other parties. The signal stays in the same expert and continues until being routed to another expert. During backpropagation, the update follows the same route as the forward pass. 
    }
    \label{fig:architecture}
\end{figure*}

We introduce \emph{privacy-preserving collaborative MoE training} (\textsc{PC-MoE}), a protocol that lets multiple parties fine-tune a \emph{single} sparse LLM while keeping both their raw data and the bulk of their model parameters strictly local. Each party hosts (i) its own backbone layers and routing networks and (ii) a small shard of the global expert pool. During training, the local router may call remote experts. Figure~\ref{fig:architecture} walks through one such step: 1. the router picks the top-$k$ experts, two of which live on remote parties; 2. only the corresponding hidden activations are sent across the network; 3. the hidden state moves to the selected expert’s host and the host runs forward pass through the expert and all subsequent local layers until another routing decision sends it further downstream; 4. gradients flow back along exactly the same sparse paths, so no full model or raw token ever leaves its silo.

Because expert activations are \emph{sparse} (top–$k$) and each party owns at most $m/n$ of the experts, any single participant observes only a vanishing fraction of another party’s forward or backward signal \cite{he2024mixture}; an adversary would need to collude with an increasingly large set of neighbours to reconstruct useful information \cite{eloul2022enhancing,dibbo2023sparse}.  At the same time, compute and memory footprints are amortized: the heavyweight expert parameters are distributed across the federation, so each party can train with dramatically smaller GPUs—our experiments show up to 70 \% peak-RAM reduction—yet the model still benefits from the \emph{full} capacity of the global expert pool, matching centralized baselines across seven LLM benchmarks while leaking virtually no content under state-of-the-art gradient-inversion attacks (\ref{Privacy}).

In short, \textsc{PC-MoE} delivers the two main promises of collaboration—larger effective models and broader data coverage—\emph{without} reviving the two hazards of hardware overload and data leakage that deter open participation today.  The remainder of the paper formalizes the protocol, analyzes its privacy guarantees and memory-footprint savings (Section~\ref{sec:methodology}), and validates its utility and scalability in extensive experiments (Section~\ref{sec:Experiments}).

\vspace{-7.5mm}

\section{Related Work}

\subsection{Collaborative Training for LLMs}
There have been efforts towards collaborative LLM training from the open-source community. \citet{borzunov2022petals} proposed a training + inference framework to allow many users to collectively run and fine-tune extremely large LLMs without each user needing access to massive hardware by distributing the shards of an LLM among the users. \citet{douillard2023diloco} proposed to use local SGD to address the communication cost of training LLMs in a distributed setting where the compute clusters are physically distant. Although each work makes valuable technical contributions, they neither address the problem that parties may lack sufficient hardware to train a full model nor tackle the privacy concerns that arise during collaboration.

\subsection{Federated Learning}
As a subfield of collaborative machine learning, federated learning was originally proposed to address the issue of training data privacy \cite{mcmahan2017communication}. Instead of sharing the training data, each client trains a model locally on their own data, and periodically sends the updated gradient to a central server. The server aggregates the gradient from all parties and broadcasts it back to each party, so that every participant will have a model updated from the gradients trained on all data. However, works such as \citet{huang2021evaluating, geiping2020inverting, hatamizadeh2023gradient} showed that, when the full gradient is shared, the adversary can exploit it with gradient inversion to reconstruct the training data. A common approach to counter it is to leverage the privacy guarantees of differential privacy by adding carefully crafted noise to the gradient before sharing \citep{wei2021gradient,wei2020federated,gauthier2023personalized}. However, it comes at a cost of model performance degradation.

\subsection{Expert Parallelization in MoE}
Expert parallelism refers to a form of model parallelism in the context of MoE, where experts are placed on different workers. Many works have proposed strategies for training MoE LLM efficiently at scale with such parallelism in mind \cite{hwang2023tutel, yao2024exploiting, pan2024parm}. In our work, we propose a form of expert parallelism that shares the experts among all parties. For this protocol, experts are the only components of the entire MoE model that are shared. In this way, each participant can harness the collective computational resources of all other participants for expert training, thereby reducing the hardware requirement (RAM in particular) on any single participant.

\subsection{Sparse Gradients as a Privacy Defense}

Recent work confirms that sparser shared information weakens gradient–inversion attacks.  \citet{eloul2022enhancing} mix multiple
examples in each gradient, while \citet{dibbo2023sparse} insert a
sparse-coding layer that transmits only a few activated coefficients. Both reduce leakage without heavy noise injection, but neither targets multi-party MoE training.  Our \textsc{PC-MoE} protocol achieves the
same effect \emph{architecturally}: top-$k$ routing means every token’s gradient touches only a handful of experts, each usually owned by a
different party, so any single adversary observes only a tiny slice of the signal.  For a survey of sparsification-based defenses, see \citet{zhang2022survey}.

\section{Methodology}
\label{sec:methodology}

In this section, we describe our proposed \emph{Privacy-preserving Collaborative MoE Training} algorithm, where multiple parties collectively train an MoE without sharing their local data, the entire model, or the full gradient, yet still benefit from the collaborative model training. We begin by discussing the overall setup, then detail the training algorithm and associated processes.

\subsection{Collaborative Setting}
\label{sec:setting}

Let $\mathcal{P}=\{P_1,\dots,P_n\}$ be the parties that jointly
fine-tune a single sparse LLM.  Each $P_i$ stores the following model components:  
(1) The \emph{local backbone} $L_i$ contains every parameter \emph{except}
the MoE experts\footnote{Note that from this point on, the term \emph{expert} denotes \emph{non-shared expert} in the standard MoE architecture, unless otherwise specified.}, and its weights as well as its gradients never leave
$P_i$.  
(2) A private shard $E_i$ of the global expert pool, obtained by
round-robin assignment so that $|E_i|=m/n$ when the model consists of $m$
experts in total. These experts remain on the
owner’s hardware but can be invoked remotely whenever another party’s
router selects them; the host then runs both the forward and backward
pass locally.  
(3) The gating layers $G_i$ (which are part of $L_i$) decide at run-time which subset of the \emph{global} experts (each potentially on a different machine) should process the current
hidden activation; every party therefore owns the entirety of the gating layers even though some of the experts it invokes may be hosted by other parties.
(4) A private dataset $D_i=\{(x,y)\}$ that never leaves the silo; raw
examples and labels are replaced by sparse activations before any data
crosses machine boundaries.  Figure \ref{fig:architecture} illustrates a
typical forward pass.

We adopt a \emph{semi-honest threat model}: during any training step a subset
$A\subseteq\mathcal{P}$ may collude and share everything they observe.
      However, the probability of forming such a coalition decays exponentially with its size. Specifically, let $a \in [0,1]$ denote the \emph{fraction} of the $n$ parties that collude 
      (so the coalition size is $|A| = a n$).  We posit an exponential‐decay prior

      \begin{equation}
  \Pr\!\bigl[\text{collusion of size }|A|=a n\bigr] = K\,\gamma^{a n}, \; 0 < \gamma < 1.
\label{eq:collusion-prob}
\end{equation}

      where $\gamma$ is a tunable \emph{decay factor} and $K$ is a normalizing constant.  
      Equivalently, if we write $p_a$ for the probability that an $a\%$ coalition forms, then
      \[
         p_a \;\propto\; \gamma^{a},
      \]
      so each additional party in the coalition multiplies the likelihood by $\gamma$.  
      For example, when $\gamma = 0.5$, a two-party collusion is only half as likely as a single-party
      breach, a three-party collusion is one-quarter as likely, and so on.  This model lets
      us analyze privacy guarantees against increasingly unlikely large-scale collusion.

\begin{table*}[t!]
\centering 
\footnotesize
\caption{Performance summary with convergence rounds, test accuracies, and per-round compute (in GFLOPs) for each method across all parties during training. Note that each method has the same GFLOPs on the same task per round. Some tasks have identical GFLOPs due to having the same training data size, batch size and max sequence length. `Acc.' for all methods refers to the converged test accuracy except No Fine-Tuning (no training involved). `Rounds' indicates the number of training rounds needed to reach that accuracy. Values after $\pm$ denote the standard deviation across models.}
\label{tab:test_acc}

\setlength{\tabcolsep}{2pt}
\begin{tabular}{@{}
  >{\raggedright\arraybackslash}m{2.5cm}@{\hspace{0.1pt}}
  c c c c c c c c
@{}}
\toprule
\multicolumn{1}{l}{Task} &
\multicolumn{1}{c}{No Fine-Tuning} &
\multicolumn{2}{c}{Isolated Baseline} &
\multicolumn{2}{c}{Centralized Baseline} &
\multicolumn{2}{c}{\textbf{PC-MoE}} &
\multicolumn{1}{c}{FLOPs / Round} \\

\cmidrule(lr){2-2}  \cmidrule(lr){3-4}  \cmidrule(lr){5-6}  \cmidrule(lr){7-8}  \cmidrule(lr){9-9}
 & Acc. & Rounds & Acc. & Rounds & Acc. & Rounds & Acc. & GFLOPs \\ 
\midrule

ARC-C                    & 15.00 & 8  & $60.70\!\pm\!0.73$ & 2 & 61.46 & \textbf{3} & $61.04\!\pm\!0.35$ & 1635.84 \\
OpenBookQA               & 53.75 & 11 & $62.12\!\pm\!1.59$ & 9 & 63.75 & 9 & $63.31\!\pm\!0.54$ & 327.17 \\
SuperGLUE                & 32.50 & 10 & $39.06\!\pm\!5.33$ & 5 & 47.50 & 9 & $\mathbf{46.25\!\pm\!2.31}$ & 1635.84 \\
MMLU-Redux               & 12.00 & 8  & $42.29\!\pm\!1.12$ & 8 & 42.33 & \textbf{3} & $41.96\!\pm\!0.86$ & 1635.84 \\
AGIEval                   & 12.50 & 10 & $24.12\!\pm\!2.08$ & 5 & 26.25 & \textbf{7} & $\mathbf{26.28\!\pm\!0.97}$ & 1635.84 \\
BBH                      &  8.60 & 7  & $11.94\!\pm\!0.83$ & 6 & 14.75 & 6 & $\mathbf{13.16\!\pm\!0.61}$ & 1635.84 \\
MedQA     & 25.20 & 9  & $26.8\!\pm\!0.6$  & 8 & 29.0  & 9 & $\mathbf{28.4\!\pm\!0.6}$  & 817.92 \\
\midrule
Average                  & 22.45 & 8.95 & $38.62\!\pm\!1.55$ & 6.01 & 40.92 & \textbf{6.47} & $\mathbf{40.28\!\pm\!0.80}$ & 1332.04 \\
\bottomrule
\end{tabular}

\end{table*}

\subsection{Training Protocol}
\label{sec:training}

Since each expert can be potentially used by any party, our collaborative training proceeds in sequential order, iterating over local data batches from different parties in an alternating fashion. The overall process is summarized in Algorithm~\ref{alg:decentralized_moe_experts_only_sequential} and Figure~\ref{fig:architecture} highlights the core idea of cross-party expert routing. Here, we highlight the key phases:

\bmhead{Forward pass} During each epoch, the parties iterate over their local mini-batches in a
round-robin schedule.  When a batch $(x_i,y_i)\!\sim\!D_i$ is up, the
hidden state starts on the data owner $P_i$ and traverses the local
backbone $L_i$ until it reaches the first routing point.  The gating
network $G_i$ selects the top–$k$ experts; for every token this yields a
set $\{E_{j_1},\dots,E_{j_k}\}$ that may reside on different hosts.
Once the activation of a token reaches a remote expert
$E_{j_\ell}\in E_{r}$, it \emph{remains on that host} $P_r$: the expert
runs its forward pass, the result feeds directly into the
\emph{next local backbone block and router on the \mbox{same} machine},
and only if that router again picks an expert on a different party is
the activation transmitted across parties, thereby reducing the inter-party communication overhead. After the final expert layer the hidden state is sent back to the originating party $P_i$ to obtain the final output and computes the loss $\mathcal{L}(h,y_i)$.

\bmhead{Backward propagation for local layers and experts} Gradients retrace the forward path in strict reverse order.
\footnote{Throughout, $\nabla_{(\cdot)}\mathcal{L}$ denotes the gradient of the loss with respect to its argument.}
\;Suppose the activation produced by a remote expert $E_{j_\ell}$ on host~$P_r$ is consumed by the very next layer $l^{\text{next}}$ on host~$P_s$.  
To update $E_{j_\ell}$ the owner $P_r$ needs only $\nabla_{h^{\text{out}}}\mathcal{L}$—the gradient with respect to the \emph{expert’s output}.  
That single tensor is transferred from $P_s$ to $P_r$; all Jacobians inside $E_{j_\ell}$ are computed locally and its parameter update never leaves $P_r$.

The same principle holds one level up: the gating network $G_i$ on the data-owner party $P_i$ requires, for each activated expert, $\nabla_{h^{\text{in}}}\mathcal{L}$, i.e., the gradient with respect to the expert’s \emph{input}.  
Each expert host therefore returns exactly one activation-sized gradient tensor to $P_i$; no party ever observes the internal weights or gradients of another.  
Consequently every host releases only the slice of information needed to service its own experts, preserving both privacy and communication efficiency.

\bmhead{No global broadcast} At the end of each iteration, there is no global synchronization. All updates happen locally for $L_i$ and at the owning party for each expert $E_j$ that was invoked.

By allowing cross-party expert calls, each party benefits from specialized experts that might be available elsewhere, thereby enhancing model capacity and performance without ever having to share raw data or private parameters or the entirty of their forward pass and gradient information. This training routine requires no global parameter synchronization: each party updates only the weights it owns—its local backbone and the experts it hosts—so the algorithm avoids any cluster-wide gradient or weight averaging. Apart from a lightweight turn-taking schedule, no further coordination is needed. Post-training inference employs the identical distributed forward pass mechanism.

\subsection{Privacy Implications}

Because no raw data or backbone parameters ever leave their owner
and each remote host sees only a \emph{vanishingly small, sparsified}
slice of the activations, the protocol offers \textbf{provably strong
privacy guarantees}:

\bmhead{Local layers remain private}
Other parties cannot infer the values of $L_i$, since no direct model parameters or updates are broadcast and only minimal gradient information (the next immediate layer) is shared.

\bmhead{Private ground truth} During the training, the ground truth (i.e., labels) of the fine-tuning data is never revealed to other parties. This guarantees the ground truth of the fine-tuning data remains entirely private through the training process.

\bmhead{Shared-expert privacy} Each router activates at most \(k\) out of the \(m\) experts, and those
experts are sharded uniformly at random across the \(n\) parties.  Hence a single adversary receives in expectation only \(k/n\) activation–gradient pairs per layer, while the remaining \(m-k\) experts—and the entire backbone—stay invisible.  Under the exponential collusion prior (Eq~\ref{eq:collusion-prob}), this yields the per-step risk bound of Eq~\ref{eq:total-risk-final}. The risk is proportional to \(\frac{k}{n}\), and is also scaled by a factor dependent on \(\gamma\) which reflects the diminishing likelihood of larger colluding coalitions, as detailed in Appendix~\ref{appendix:derivation} where the approximation for \(K\) under large \(n\) is also discussed.
Concretely, with \(k=2\) and \(n=8\), the reduction factor compared to a fully shared model is approximately \( \frac{n(1-\gamma)}{k\gamma} = 4 \frac{1-\gamma}{\gamma}\) (leveraging the large-\(n\) approximation for \(K\) detailed in the Appendix). For \(\gamma=0.5\), this yields a 4-fold reduction. To achieve a 10-fold reduction, \(\gamma\) would need to be approximately 0.28 or less. In the paragraph on empirical privacy validation in Section~\ref{sec:Experiments}, we empirically validate our privacy claim and show even in the worst-case scenario, virtually no useful signal from training data can be extracted under our training approach.

\subsection{Amortizing Per-Party Computational Resource Requirements}

Since the experts are distributed across all $n$ parties, assuming top-$k$ experts activation and each expert having an equal chance of being activated, for each party, our proposed training method on average only costs $\frac{k}{n}$ of the RAM used by expert layers a vanilla MoE training approach uses in training expert layers while the rest of the RAM usage is distributed among other parties who own the activated experts. Since expert parameters constitute the majority of the trainable parameters in an MoE LLM ($93.11\%$ in our experiments), given the same hyperparameter configuration, this results in a substantial amount of GPU RAM reduction from each individual party's perspective. As $n$ increases, the benefit of reduction becomes even more prominent. In the paragraph on memory efficiency in Section~\ref{Memory}, we show in practice the peak GPU memory usage reduction achieved is indeed substantial.

 Overall, we now have an efficient yet privacy-preserving MoE training method with minimal risk on data leakage and lower GPU RAM requirement during training.

\begin{table*}[t!]
  \centering
  \caption{Peak GPU RAM usage (MB) averaged across parties}
  \label{tab:ram}
  \small
  \setlength{\tabcolsep}{6pt}
  \renewcommand{\arraystretch}{1.05}

  \begin{tabular}{l c c c S S}
    \toprule
    & \multicolumn{3}{c}{Experts Params} & & \\[-0.4em]
    \cmidrule(lr){2-4}
    \textbf{Task}
      & {Isolated}
      & {Centralised}
      & {\textbf{PC-MoE}}
      & {Non-expert Params}
      &  {Relative Total RAM\,(\%)}\\
    \midrule
    ARC-C        & 25691.11 & 25684.34 & 2864.24 &   2194.93 & \textbf{18.14}\\
    OpenBookQA   & 25648.81 & 25635.64 & 2939.48 &  1913.78 & \textbf{17.61}\\
    SuperGLUE    & 27693.28 & 27693.77 & 7538.13 &  7908.05 & 43.39 \\
    MMLU-Redux   & 25894.59 & 25898.92 & 5382.50 &  5371.74 & \textbf{34.40} \\
    AGIEval      & 28757.78 & 28764.09 & 9211.45 &  9912.30 & 49.45 \\
    BBH          & 25835.52 & 25830.85 & 5700.03 &  5743.79 & \textbf{36.24} \\
    MedQA        & 25836.48 & 25837.18 & 4991.80 &  5891.05 & \textbf{34.30} \\
    \midrule
    \textbf{Average}
                 & 26412.40 & 26410.05 & 5342.38 & 5339.07 &  \textbf{33.64}\\
    \bottomrule
  \end{tabular}

\footnotesize\textit{Note:} \emph{Relative Total RAM} (\%) is calculated as
$\displaystyle
\frac{\text{Ours}+\text{Non-expert Params}}{\text{Isolated}+\text{Non-expert Params}}
\times 100$; smaller values indicate better memory efficiency.
\vspace{-3mm}
\end{table*}

\begin{table*}[!ht]
\small
\centering
\caption{Partial-gradient attack results on \textsc{OpenBookQA}, \textsc{AGIEval} and \textsc{ARC-C}
(10 reconstructions, 1000 optimization steps each).  
“ROUGE-1+2” is the sum of ROUGE-1 and ROUGE-2 $F_1$ scores.  
Across all datasets, the attacker recovers only a tiny fraction of
unigrams, almost no longer subsequences (ROUGE-L/L$_\text{sum}\!\approx\!$ ROUGE-1), and none of the bigrams, showing that virtually no meaningful
content is leaked.}
\label{tab:rouge-agg}

\setlength{\tabcolsep}{4pt}
\renewcommand{\arraystretch}{0.95}

\begin{tabular}{lccccc}
\toprule
            & \multicolumn{5}{c}{$F_1$ (\%)} \\ \cmidrule(lr){2-6}
\textbf{Dataset} & \textbf{ROUGE-1} & \textbf{ROUGE-2} & \textbf{ROUGE-1+2} & \textbf{ROUGE-L} & \textbf{ROUGE-L$_\text{sum}$} \\
\midrule
OpenBookQA & 7.647 & 0.000 & 7.647 & 7.679 & 7.655 \\
AGIEval    & 2.286 & 0.000 & 2.286 & 2.311 & 2.298 \\
ARC-C      & 4.670 & 0.000 & 4.670 & 4.650 & 4.663 \\
\bottomrule
\end{tabular}
\end{table*}

\begin{table*}[t!]
\small
\setlength{\tabcolsep}{1pt}
\renewcommand{\arraystretch}{0.85}
\centering
\caption{Ablation study on the effect of skipping the sharing of the first $n_s$, the last $n_s$, and both the first and last $n_s$ expert layers on test accuracy. It can be observed that skipping a few $n_s$ expert layers can sometimes be marginally beneficial for test accuracy, but skipping more layers in general will result in worse test accuracy, worse training stability (i.e., larger standard deviation) and eventually converging to the isolated baseline case.  
Isolated baseline corresponds to \(n_s{=}16\) (i.e., 16 expert layers in total and no expert layer is shared); \textsc{PC-MoE} corresponds to \(n_s{=}0\) (i.e., all expert layers are shared). For Skip First+Last $n_s$ column, $n_s$'s value is the total number of skipped layer, e.g., $n_s=8$ means skipping first 4 and last 4 layers.}
\label{tab:skip_ablation_compact}

\begin{tabular}{@{}l l
                c c  
                c c  
                c c  
                @{}}
\toprule
\multirow{2}{*}{Task} & \multirow{2}{*}{\(n_s\)} &
\multicolumn{2}{c}{Skip First \(n_s\)} &
\multicolumn{2}{c}{Skip Last \(n_s\)} &
\multicolumn{2}{c}{Skip First\,+\,Last \(n_s\)} \\ 
\cmidrule(lr){3-4}\cmidrule(lr){5-6}\cmidrule(lr){7-8}
 & & Rounds & Acc. & Rounds & Acc. & Rounds & Acc. \\ 
\midrule
\multirow{7}{*}{ARC-C}
 & 0  (\textbf{PC-MoE})             & \textbf{3} & $61.04\!\pm\!0.35$ & \textbf{3} & $61.04\!\pm\!0.35$ & \textbf{3} & $61.04\!\pm\!0.35$ \\ 
 & 2                     & 4 & $61.04\!\pm\!0.55$                 & 3 & $60.89\!\pm\!0.33$                 & 4 & $60.76\!\pm\!0.43$ \\
 & 4                     & 5 & $60.68\!\pm\!0.44$                 & 4 & $61.12\!\pm\!0.40$                 & 4 & $61.07\!\pm\!0.17$ \\
 & 8                     & 6 & $60.76\!\pm\!0.25$                 & 4 & $61.02\!\pm\!0.26$                 & 5 & $60.65\!\pm\!0.26$ \\ 
 & 12                    & 7 & $60.23\!\pm\!0.17$                  & 7 & $61.43\!\pm\!0.21$                 & 6 & $60.96\!\pm\!0.27$ \\
 & 14                    & 7 & $60.99\!\pm\!0.59$                 & 7 & $60.37\!\pm\!0.50$                 & 8 & $60.89\!\pm\!0.50$ \\
 & 16 (Isolated Baseline)         & 8  & $60.70\!\pm\!0.73$ & 8  & $60.70\!\pm\!0.73$ & 8  & $60.70\!\pm\!0.73$ \\ 
\midrule

\multirow{7}{*}{SuperGLUE}
 & 0  (\textbf{PC-MoE})             & 9  & $\mathbf{46.25\!\pm\!2.31}$ & 9  & $\mathbf{46.25\!\pm\!2.31}$ & 9  & $46.25\!\pm\!2.31$ \\
 & 2                     & 7 & $43.59\!\pm\!2.45$                 & 9 & $44.69 \!\pm\!2.19$                & 11 & $44.22 \!\pm\!1.63$  \\
 & 4                     & 8 & $45.00 \!\pm\!2.31$                  & 10 & $45.16 \!\pm\!1.24$                  & 11 & $44.84 \!\pm\!1.70$ \\
 & 8                     & 8 & $45.00 \!\pm\!6.01$                 & 10 & $42.34 \!\pm\!2.54$                  & 11  & $\mathbf{46.41 \!\pm\!1.04}$ \\ 
 & 12                    & 11 & $37.66 \!\pm\!7.48$                 & 10 & $41.88 \!\pm\!2.91$                  & 11 & $42.81 \!\pm\!2.48$ \\
 & 14                    & 11 & $38.91 \!\pm\!5.24$                 & 10 & $39.69 \!\pm\!3.04$                  & 11 & $38.91 \!\pm\!5.49$ \\
 & 16 (Isolated Baseline)         & 10 & $39.06\!\pm\!5.33$ & 10 & $39.06\!\pm\!5.33$ & 10 & $39.06\!\pm\!5.33$ \\
\midrule
\multirow{7}{*}{AGIEval}
 & 0 (\textbf{PC-MoE})             & \textbf{7} & $\mathbf{26.28\!\pm\!0.97}$ & \textbf{7} & $\mathbf{26.28\!\pm\!0.97}$ & \textbf{7} & $\mathbf{26.28\!\pm\!0.97}$ \\
 & 2                     & 9 & $26.01\!\pm\!0.75$                 & 9 & $25.38\!\pm\!0.80$                 & 8 & $25.66\!\pm\!0.83$ \\
 & 4                     & 10 & $25.98\!\pm\!0.82$                 & 9 & $25.38\!\pm\!0.98$                 & 10 & $25.81\!\pm\!0.93$ \\
 & 8                     & 9 & $25.38\!\pm\!1.03$                  & 8 & $23.99\!\pm\!2.31$                 & 7 & $24.91\!\pm\!1.41$ \\
 & 12                    & 10 & $24.87\!\pm\!1.22$                  & 10 & $23.54\!\pm\!1.88$                  & 10 & $25.52\!\pm\!1.70$ \\
 & 14                    & 10 & $24.62\!\pm\!1.11$                  & 10 & $24.69\!\pm\!1.73$                  & 10 & $25.19\!\pm\!0.36$ \\
 & 16  (Isolated Baseline)         & 10 & $24.12\!\pm\!2.08$ & 10 & $24.12\!\pm\!2.08$ & 10 & $24.12\!\pm\!2.08$ \\
\midrule
\multirow{7}{*}{\textbf{Average}}
 & 0 (\textbf{PC-MoE})             & \textbf{5.76
} & $\mathbf{45.47 \!\pm\!1.01}$ & \textbf{5.76
} & $\mathbf{45.47 \!\pm\!1.01}$ & \textbf{5.76
} & $\mathbf{45.47 \!\pm\!1.01}$ \\
 & 2                     & 6.44
   & $44.77\!\pm\!1.30$                 & 6.46   & $44.74 \!\pm\!0.91$                 & 6.99   &$44.67 \!\pm\!0.84$
 \\
 & 4                     & 7.44
   & $44.93\!\pm\!1.23$                  & 7.11
   & $44.94 \!\pm\!0.79$                 & 7.69
   & $45.00  \!\pm\!0.78$ \\
 & 8                     & 7.51
   & $44.67\!\pm\!2.93$                  & 6.76
   & $43.77 \!\pm\!1.50$                 & 7.06 & $44.86 \!\pm\!0.84$ \\
 & 12                    & 8.54
   & $42.69\!\pm\!3.63$                  & 8.73   & $43.68 \!\pm\!1.41$                 & 8.54   &  $44.39 \!\pm\!1.27$\\
 & 14                    & 8.96
   & $43.21\!\pm\!2.60$                  & 8.73   & $43.40 \!\pm\!1.50$                 & 9.38   &  $43.37 \!\pm\!1.58$\\
 & 16  (Isolated Baseline)         & 9.15
 & $42.94
\!\pm\!2.24
$ & 9.15
 & $42.94
\!\pm\!2.24
$ & 9.15
 & $42.94
\!\pm\!2.24
$ \\
\bottomrule
\end{tabular}
\end{table*}

\section{Experiments}
\label{sec:Experiments}
\bmhead{Setup} We compared our proposed collaborative MoE LLM training with two baselines: when the parties train models only on their own data (i.e., the isolated baseline) and when they share their training data to train a single model (i.e., the centralized baseline). In our approach, since a single expert may be required by multiple parties, we schedule training in alternating blocks. Specifically, party $i$ trains for $B$ steps, then hands off to party $j$, and the process continues in this round-robin fashion. The sequence of parties is randomized at the start of each cycle, and every party receives exactly one training turn before the cycle repeats.

\bmhead{Performance} We test our training method on the following popular benchmarks: ARC-C \cite{clark2018think}, OpenBookQA \cite{mihaylov2018can}, SuperGLUE \cite{sarlin2020superglue}, MMLU-Redux \cite{gema2024we}, AGIEval \cite{zhong2023agieval}, BIG-Bench Hard (BBH) \cite{suzgun2022challenging} and MedQA \cite{jin2021disease} with eight parties as the default setting and three exemplars for few-shot prompting. For hardware, we train all the models on H100 GPU. As shown in Table~\ref{tab:test_acc}, our method on average has a converged test accuracy and convergence rate almost matches that of the centralized setting where all data are collected to train a single model.  This trend is consistent across various tasks we have tested on \footnote{Note that some of the improvements compared to the base model are marginal because, in general, it is difficult to achieve substantial gains on common LLM benchmarks through fine-tuning with very limited data (i.e., pre-training is more crucial for such improvements). However, the relative improvement of our method compared to the baselines is significant.}. Details on each task training can be found in Appendix~\ref{appendix:Experiment}.

\label{Memory}
\bmhead{Memory efficiency} Table~\ref{tab:ram} shows the peak GPU RAM usage, as well as how much memory our method saved w.r.t. the isolated baseline on the same benchmarks. Overall, our method achieved  around 70\% peak RAM usage reduction compared to the isolated baseline (centralized baseline has negligible RAM usage difference compared to the isolated case) for each party. This means individual party can now conduct MoE LLM training on the same scale on much smaller GPUs, greatly lowering the hardware barrier to entry.

\begin{figure*}[!t]
	\centering
	\subfloat[2-party setup]{\label{fig:openbookqa_cycle_accuracy_comparison_ablation_2parties}
	   \includegraphics[width=0.3\linewidth]{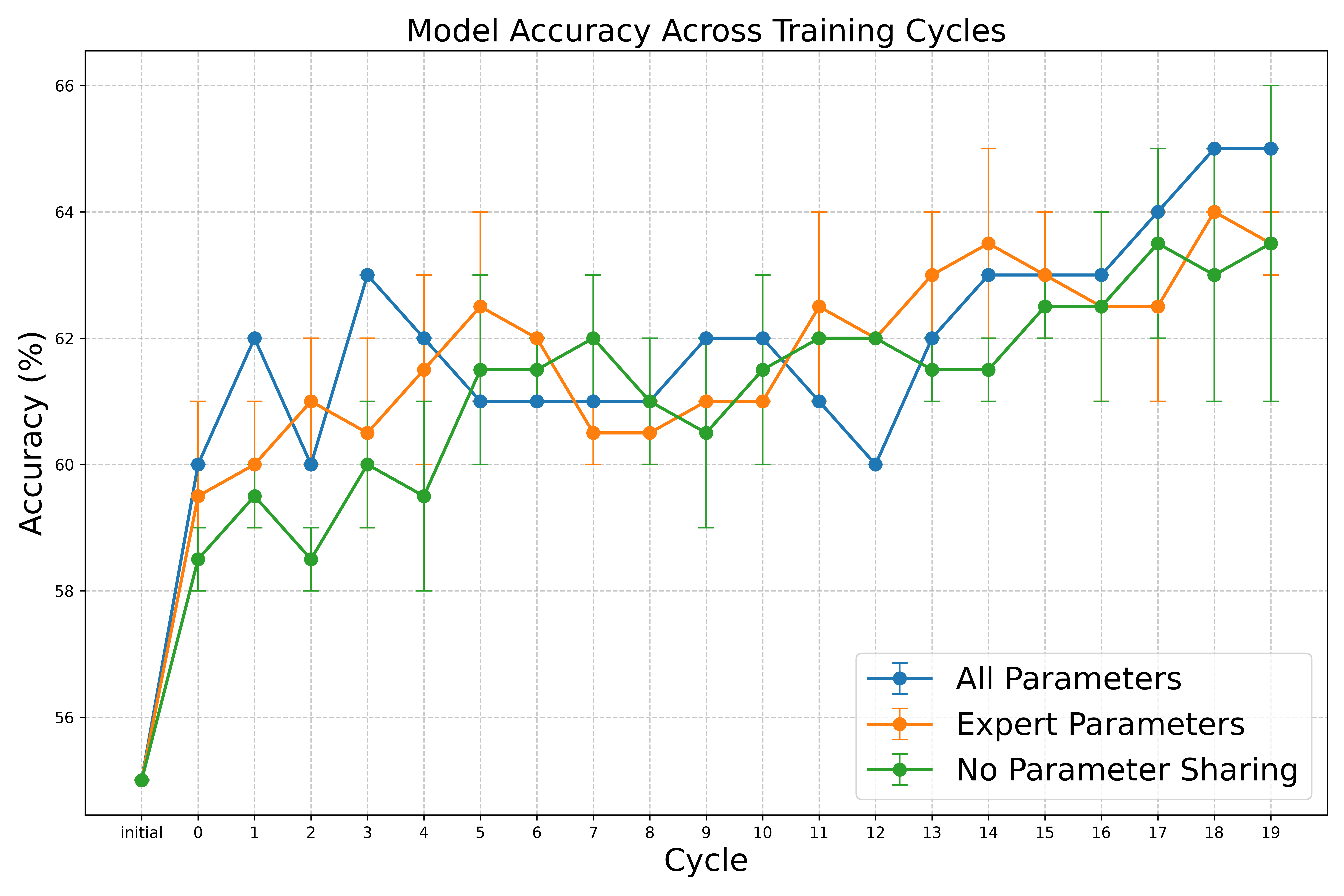}}
    ~~\subfloat[4-party setup]{\label{fig:openbookqa_cycle_accuracy_comparison_ablation_4parties}
		\includegraphics[width=0.3\linewidth]{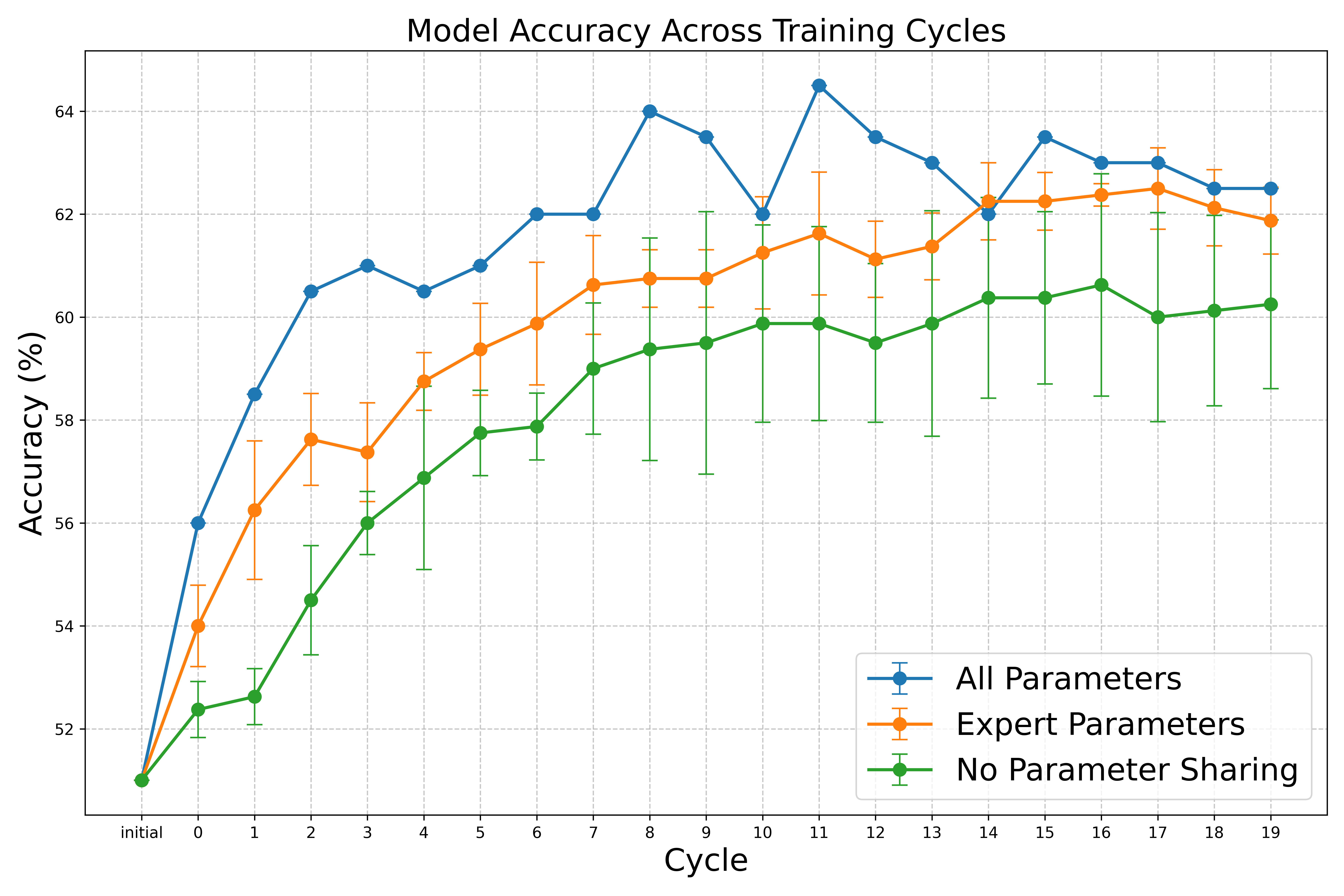}}
    ~~\subfloat[8-party setup]{\label{fig:openbookqa_cycle_accuracy_comparison}
		\includegraphics[width=0.3\linewidth]{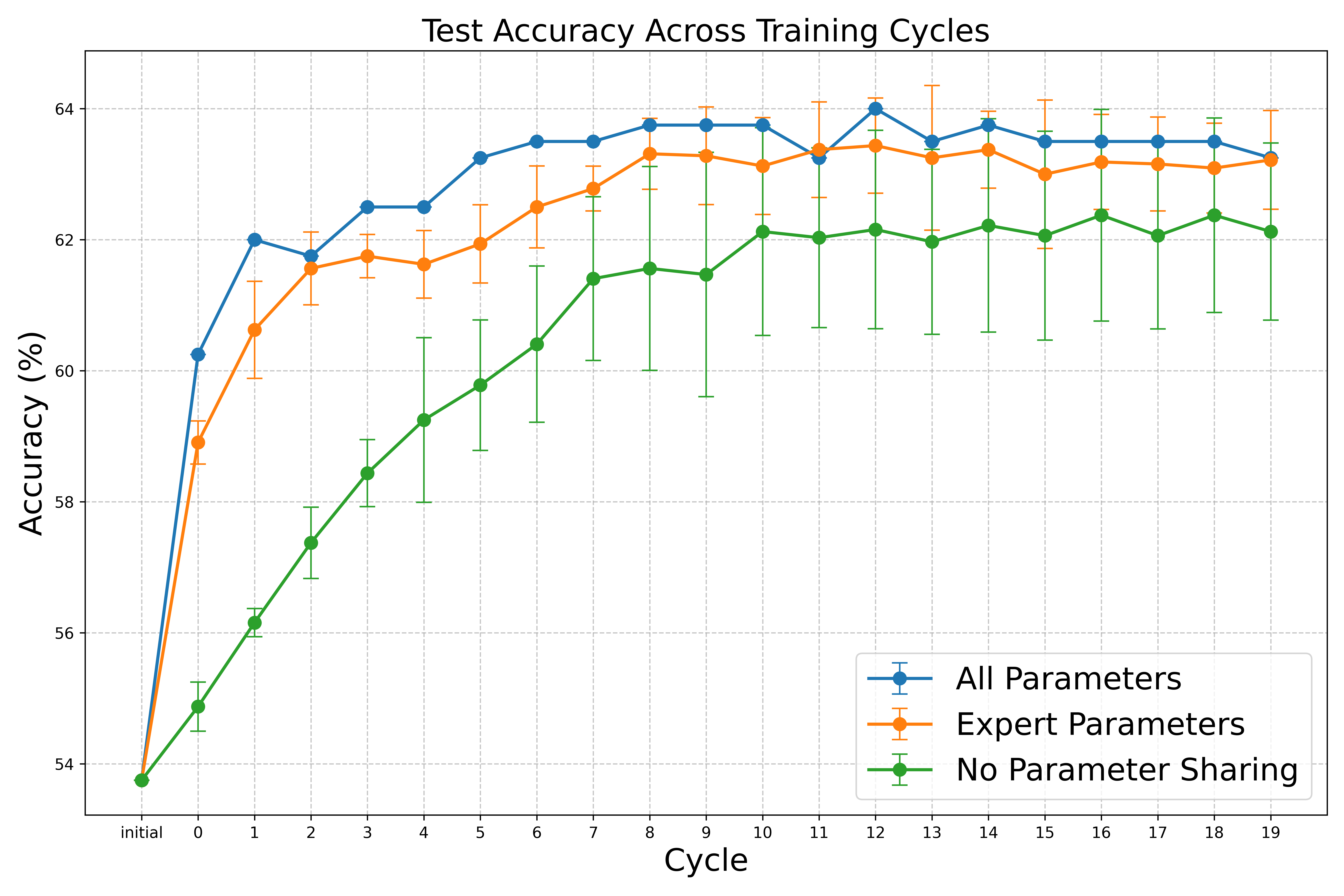}}
	\vspace{-1mm}
    \caption{ 
        Ablation on the scalability of our method on OpenBookQA. As shown, when each party's data remains the same, with more number of parties we observed more significant difference and benefit in terms of converged test accuracy when using our distributed training method.
    }
	\label{fig:ablations_scalability1}
    \vspace{-3.5mm}
\end{figure*}

\label{Privacy}
\bmhead{Empirical validation on privacy guarantee} Since the first gradient-inversion attacks were proposed, subsequent work has aimed to relax the assumptions required to reconstruct training data. \cite{petrov2024dager} proposed a reconstruction attack that exactly reconstructs whole batches of input text from just the self-attention gradients of large language models, exploiting their low-rank structure and discrete token embeddings. The attack proposed by \cite{li2024seeing} is even less restrictive, as it does not require the partial gradient from the self-attention layer. Since our method only shares each party's expert layers, we use the partial gradient attack introduced by \cite{li2024seeing} to validate our privacy guarantee\footnote{In our setup, both the data and labels are privately owned by each party and remain in-silo throughout the collaboration, membership inference attack is ineffective \cite{shokri2017membership}.}. To simulate a worst-case attack, we give the adversary full access to the gradients of every expert layer. Note that in this setting, the attack has a higher chance of success than would occur in practice and our threat model anticipates. As Table~\ref{tab:rouge-agg} shows, empirically we observe that the partial-gradient attack is not able to reconstruct the training data OpenBookQA, AGIEval and ARC-C to any recognizable extent: across ten questions, the attack reproduces only around 8 \%, 2 \% and 5 \% of the reference unigrams, almost no longer subsequences (ROUGE-L/L$_\text{sum} \approx$ ROUGE-1), and none of the bigrams for them respectively, indicating that virtually no meaningful content is recovered. These empirical findings confirm our earlier theoretical analysis of the privacy guarantees offered by our training approach.

\subsection{Ablation}

\bmhead{Effect of the number of shared expert layers} To further understand how the amount of expert parameters shared impacts the model performance, we conduct the following ablation experiments. We observe the test accuracy when the parties share only the first $n_s$ expert layers, the last $n_s$ expert layers, and both the first $n_s$ and the last $n_s$ layer respectively. As shown in Table~\ref{tab:skip_ablation_compact}, in some cases, skipping a few expert layers can be beneficial for model performance which is likely due to less overfitting. However, in general, skipping more layers will result in worse test accuracy or slower convergence, making training more unstable (larger standard deviation) than sharing all expert layers, and converging to the performance of isolated baseline case. Since we have shown that sharing all the expert layers does not result in privacy breach, it is not necessary to skip sharing expert layers in practice to keep the benefit on GPU RAM reduction.

\bmhead{Scalability} It is also crucial to investigate if and how the number of parties affects the performance of the proposed method. To this end, we fine-tune on three selected tasks with two and four parties respectively. As shown in Figures~\ref{fig:ablations_scalability1} and \ref{fig:ablations_scalability2}, the performance benefits (convergence rate and converged test accuracy) of our method become more prominent as the number of parties increases. This result indicates that our method scales effectively as the number of parties increases.

\section{Conclusion}
In this work, we propose the Collaborative MoE Training protocol \textsc{PC-MoE}, which enables multiple parties to leverage each other's compute resources and data to reduce the hardware burden and improve their own MoE LLM's performance without sacrificing data privacy. It demonstrates that preserving privacy does not always have to come at a steep cost to utility or vice versa: by routing only sparse expert signals, parties obtain near-centralized performance, enjoy lower hardware requirements, while revealing virtually nothing to an attacker. We hope our results encourage future privacy research to explore how far the utility frontier can be pushed with minimal compromise in privacy and safety.

\begin{appendices}

\section{More Experiment Details}
\label{appendix:Experiment}
\subsection{Training}
\label{appendix:Training}

\begin{figure*}[!ht]
	\centering
	\subfloat[agieval]{\label{fig:agieval_cycle_accuracy_comparison}
	   \includegraphics[width=0.31\linewidth]{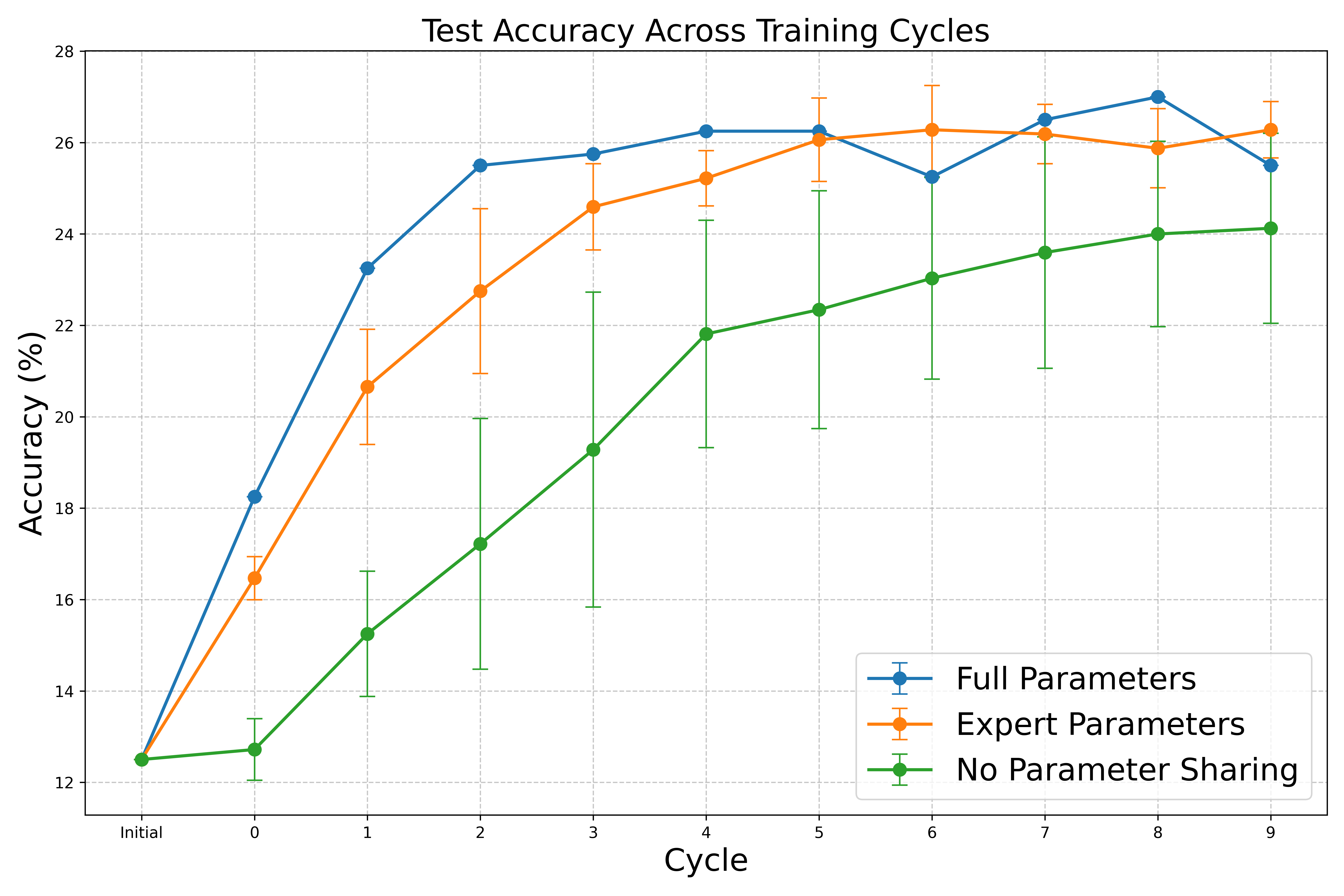}}
    ~~\subfloat[ARC]{\label{fig:arc_cycle_accuracy_comparison}
		\includegraphics[width=0.31\linewidth]{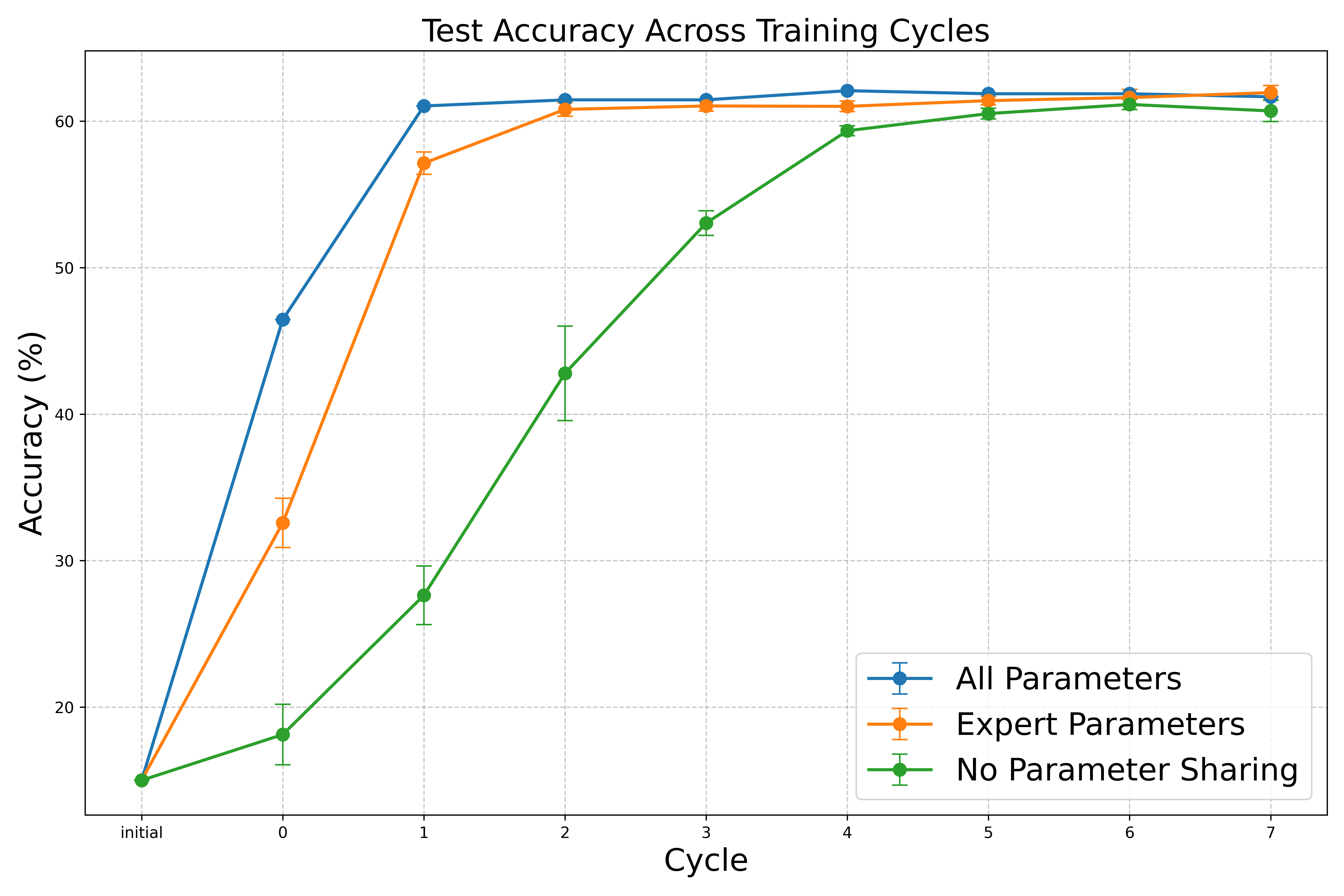}}
	~~\subfloat[BIG-Bench Hard]{\label{fig:bbh_cycle_accuracy_comparison}
		\includegraphics[width=0.31\linewidth]{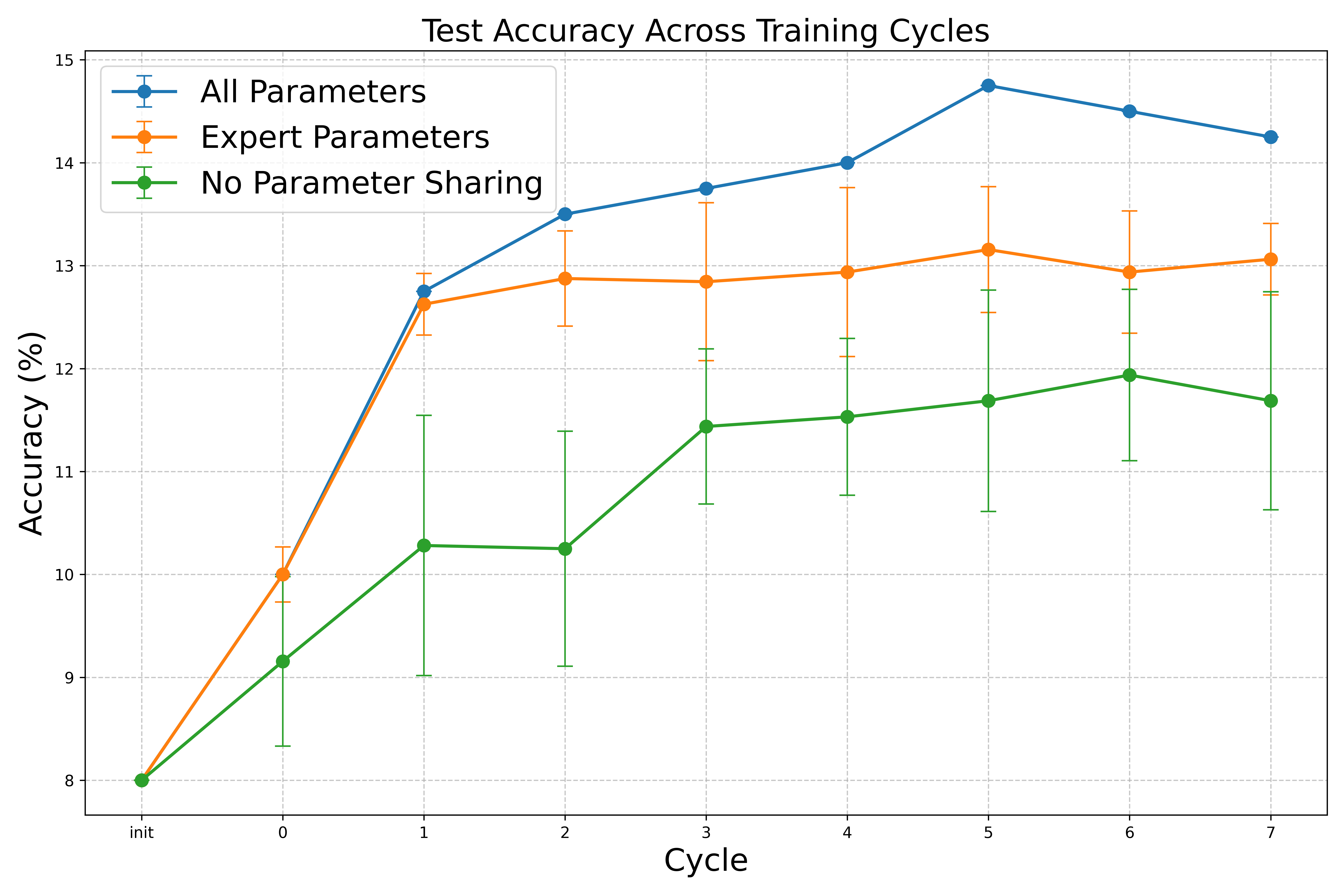}} \\
\subfloat[MedQA]{\label{fig:medqa_cycle_accuracy_comparison}
	   \includegraphics[width=0.25\linewidth]{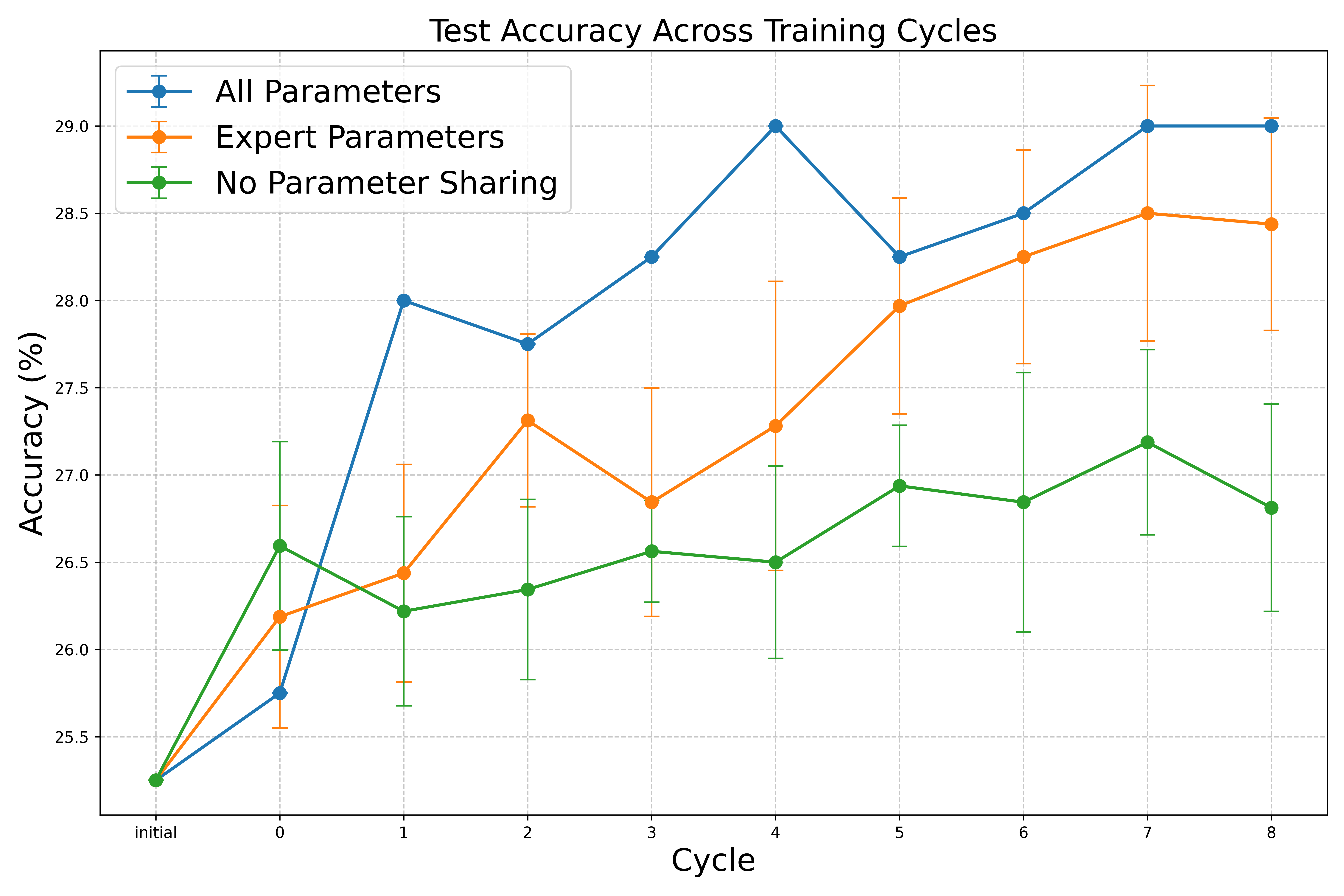}}
    ~~\subfloat[MMLU-Redux]{\label{fig:mmlu_redux_cycle_accuracy_comparison}
		\includegraphics[width=0.25\linewidth]{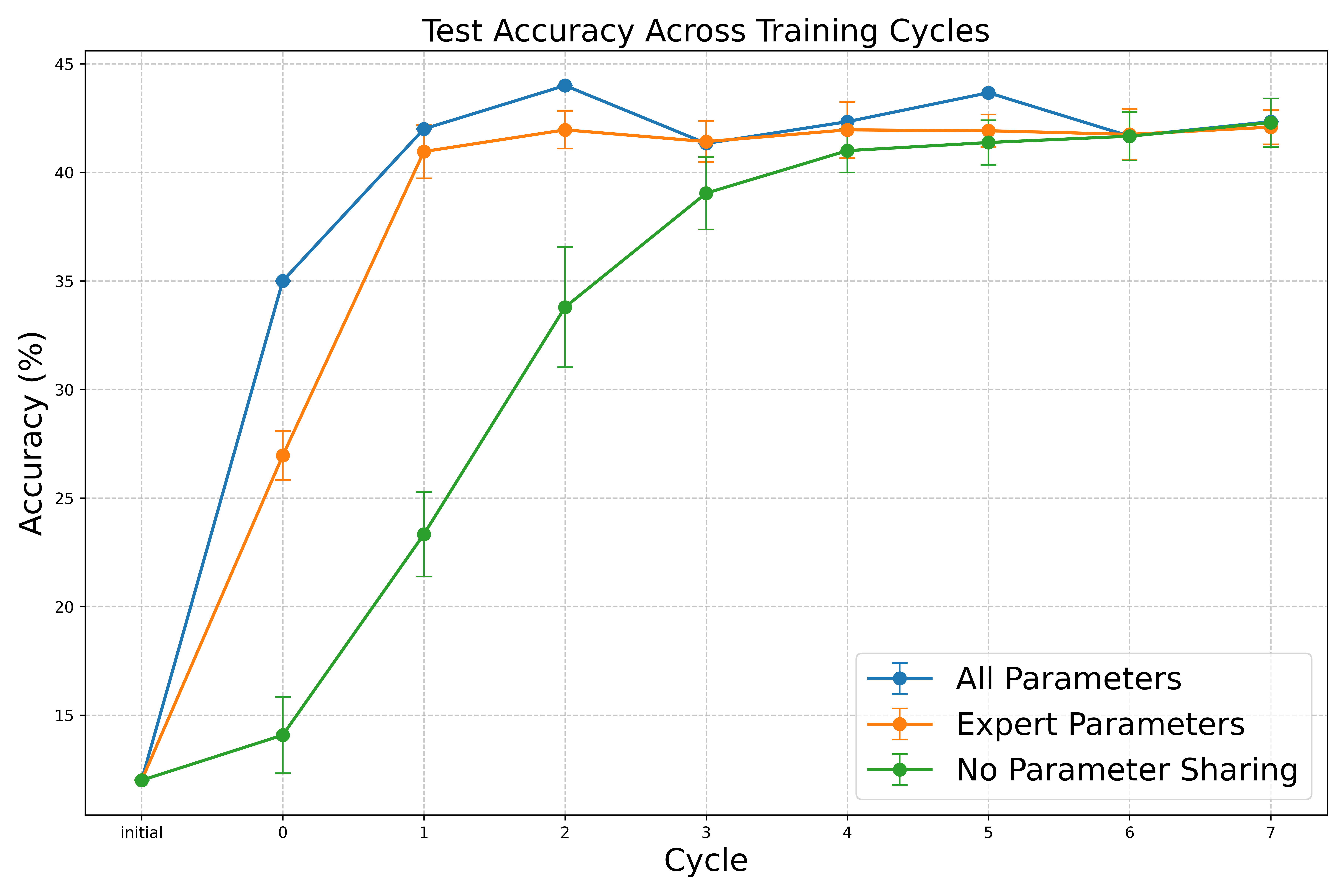}}
	~~\subfloat[OpenBookQA]{\label{fig:openbookqa_cycle_accuracy_comparison2}
	\includegraphics[width=0.25\linewidth]{openbookqa_cycle_accuracy_comparison.png}}
    ~~\subfloat[SuperGLUE]{\label{fig:superglue_cycle_accuracy_comparison}
		\includegraphics[width=0.25\linewidth]{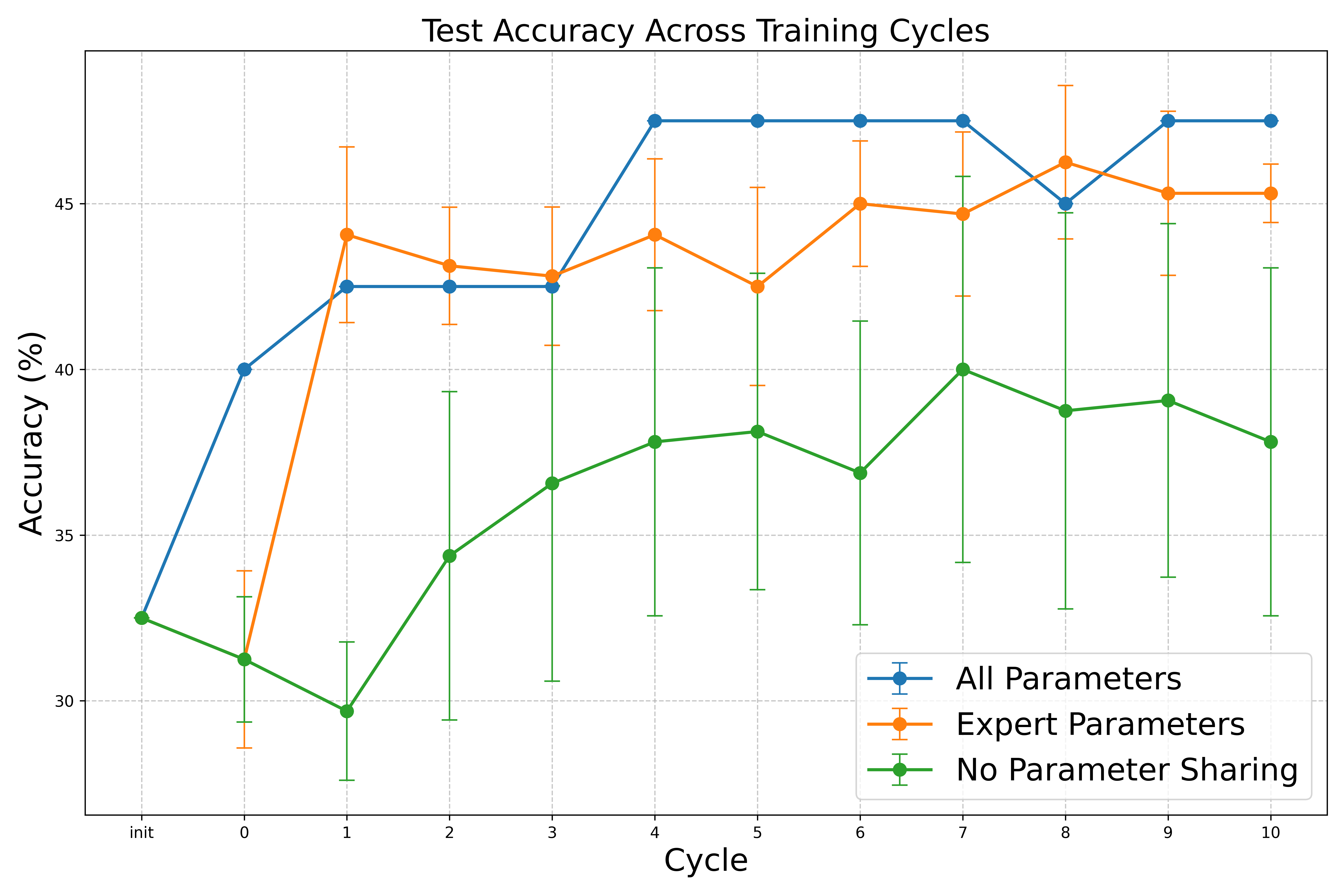}}
	\vspace{-1mm}
    \caption{ 
        Test accuracy plots of the tasks in Table~\ref{tab:test_acc} during training. The error bars indicate standard deviation. From the plots, it can be more clearly seen that our method either provides a better converged accuracy that is close to the centralized baseline or it has fast convergence rate that is close to the centralized baseline (or both) compared to the isolated baseline.
    }
	\label{fig:ablations77}
    \vspace{-3.5mm}
\end{figure*}

Figure~\ref{fig:ablations77} shows the test accuracy plots of the tasks in Table~\ref{tab:test_acc} during training. It can be observed that \textsc{PC-MoE} has similar test accuracy to centralized baseline as compared to isolated baseline throughout the training.

\subsection{Scalability}
\begin{figure*}[!ht]
	\centering
	\subfloat[2-party setup]{\label{fig:agieval_cycle_accuracy_comparison_2parties}
	   \includegraphics[width=0.32\linewidth]{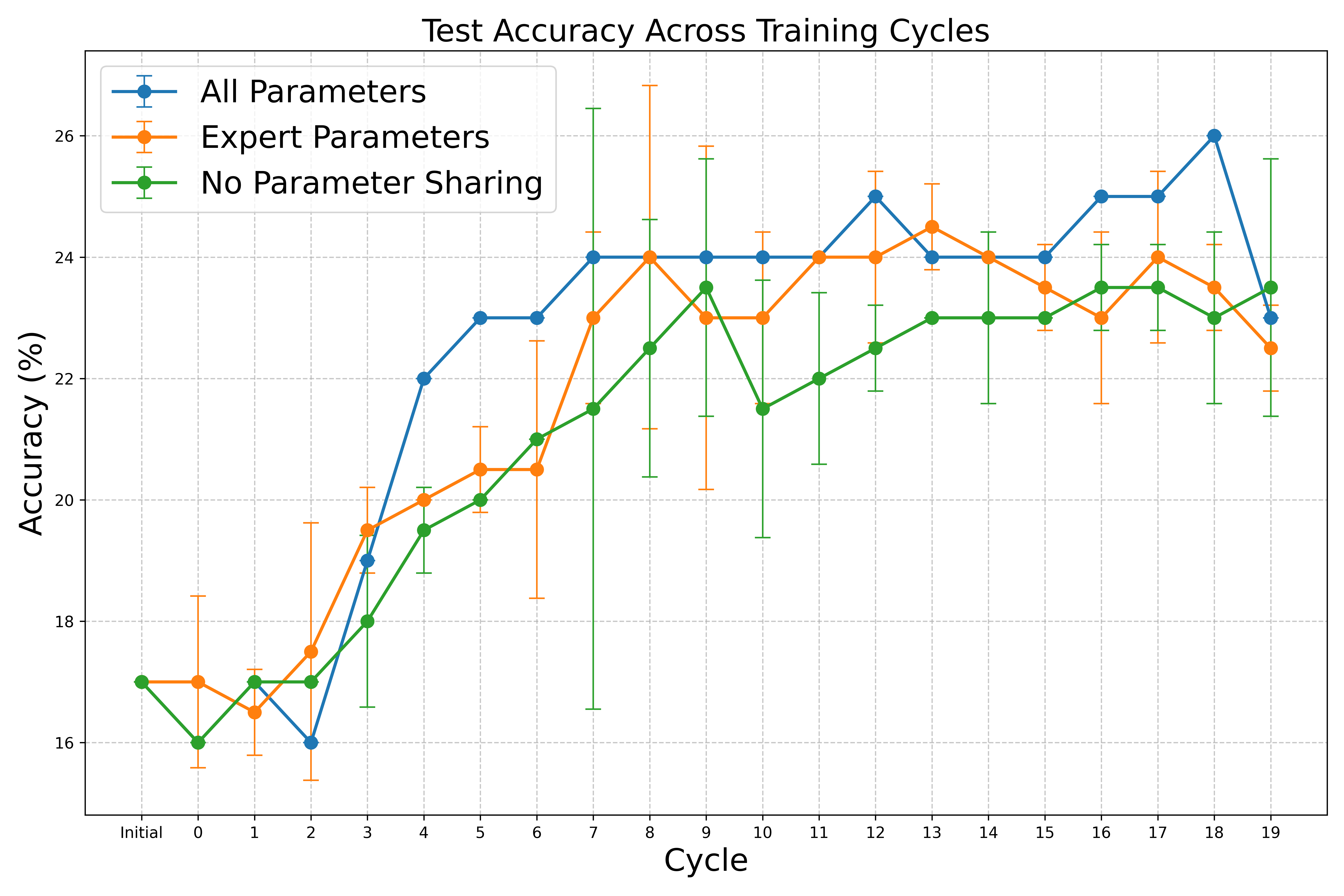}}
    ~~\subfloat[4-party setup]{\label{fig:agieval_cycle_accuracy_comparison_4parties}
		\includegraphics[width=0.32\linewidth]{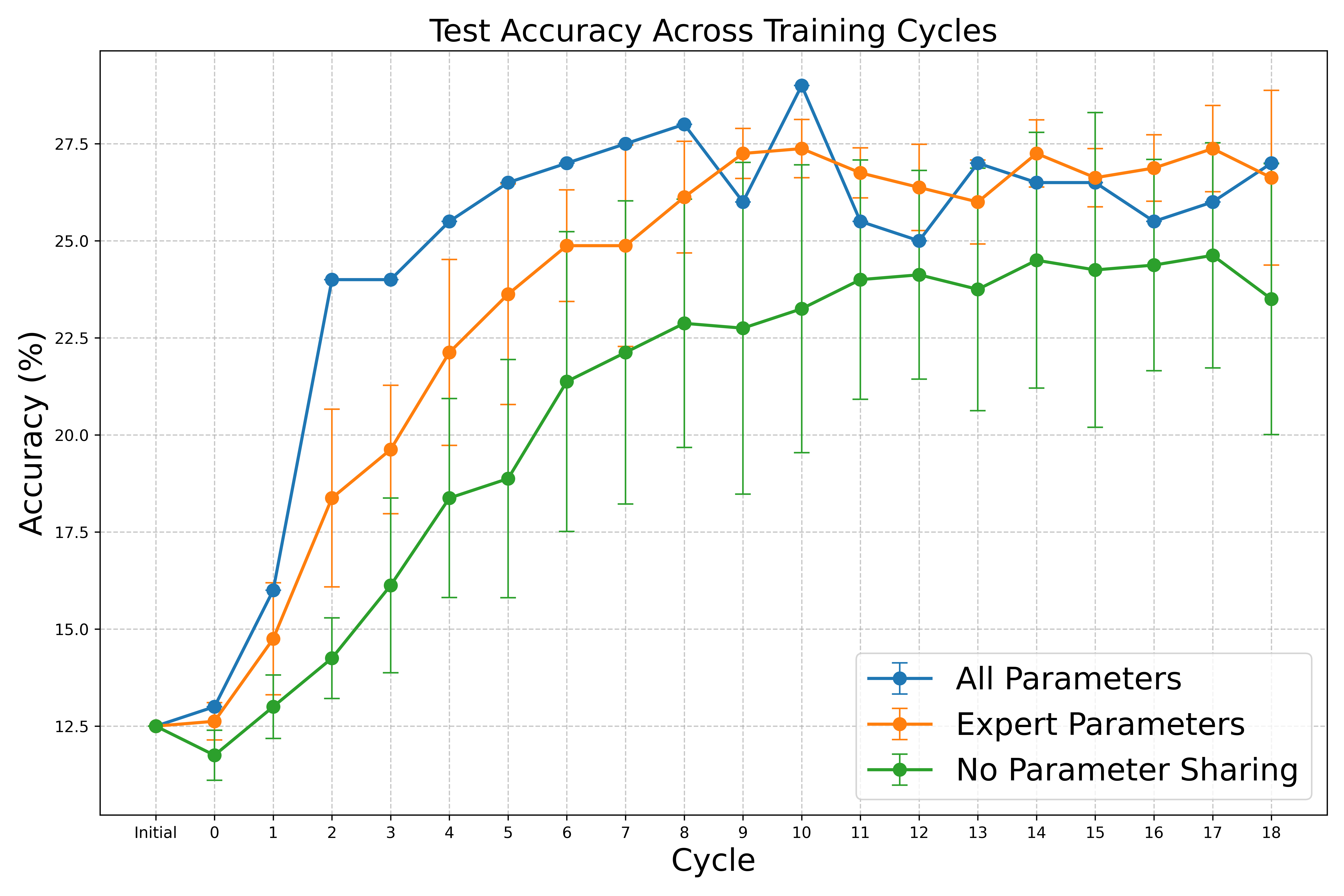}}
    ~~\subfloat[8-party setup]{\label{fig:agieval_cycle_accuracy_comparison2}
		\includegraphics[width=0.32\linewidth]{agieval_cycle_accuracy_comparison.png}}\\
        \subfloat[2-party setup]
        {\label{fig:arc_cycle_accuracy_comparison_2parties}
	   \includegraphics[width=0.32\linewidth]{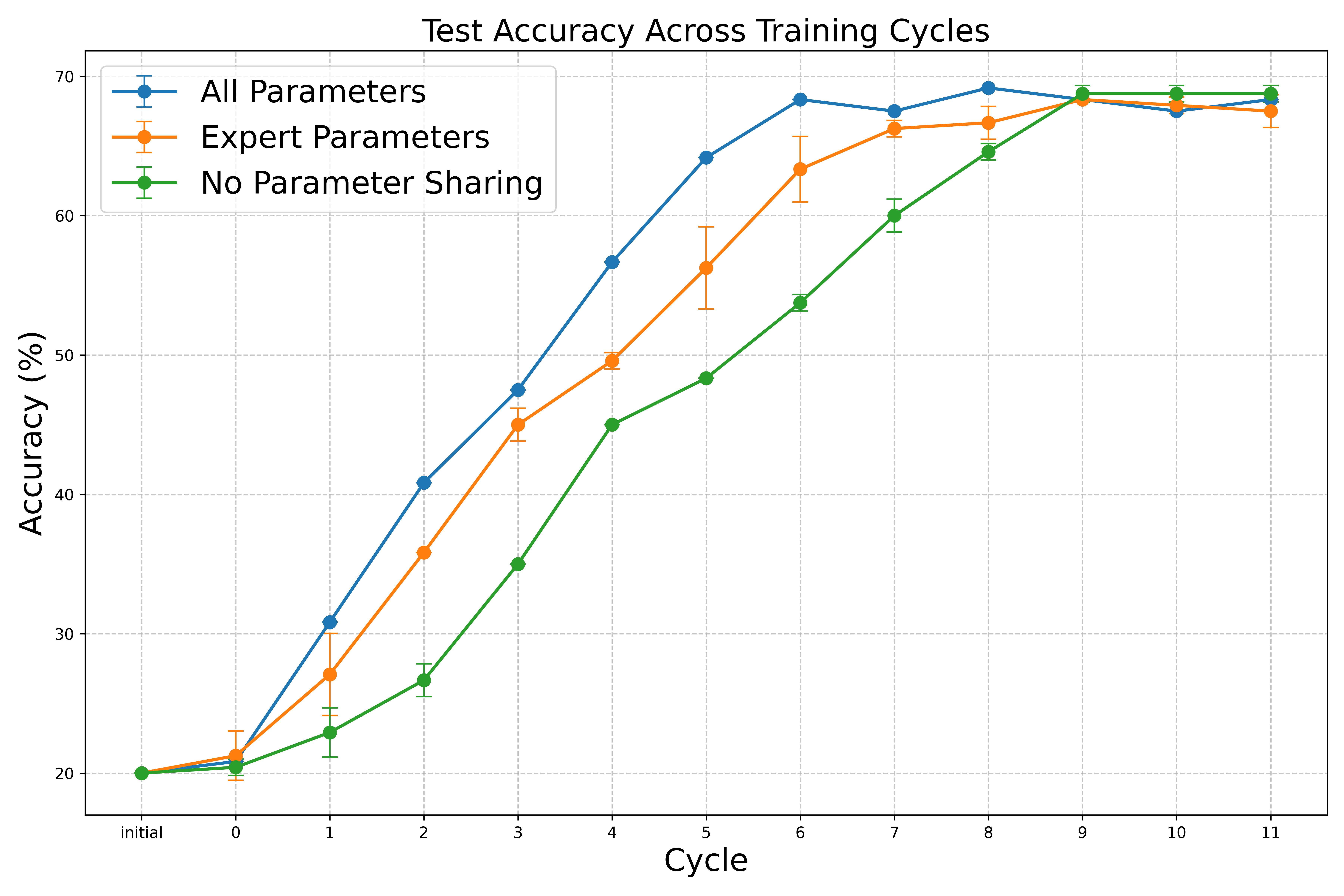}}
    ~~\subfloat[4-party setup]{\label{fig:arc_cycle_accuracy_comparison_4parties}
		\includegraphics[width=0.32\linewidth]{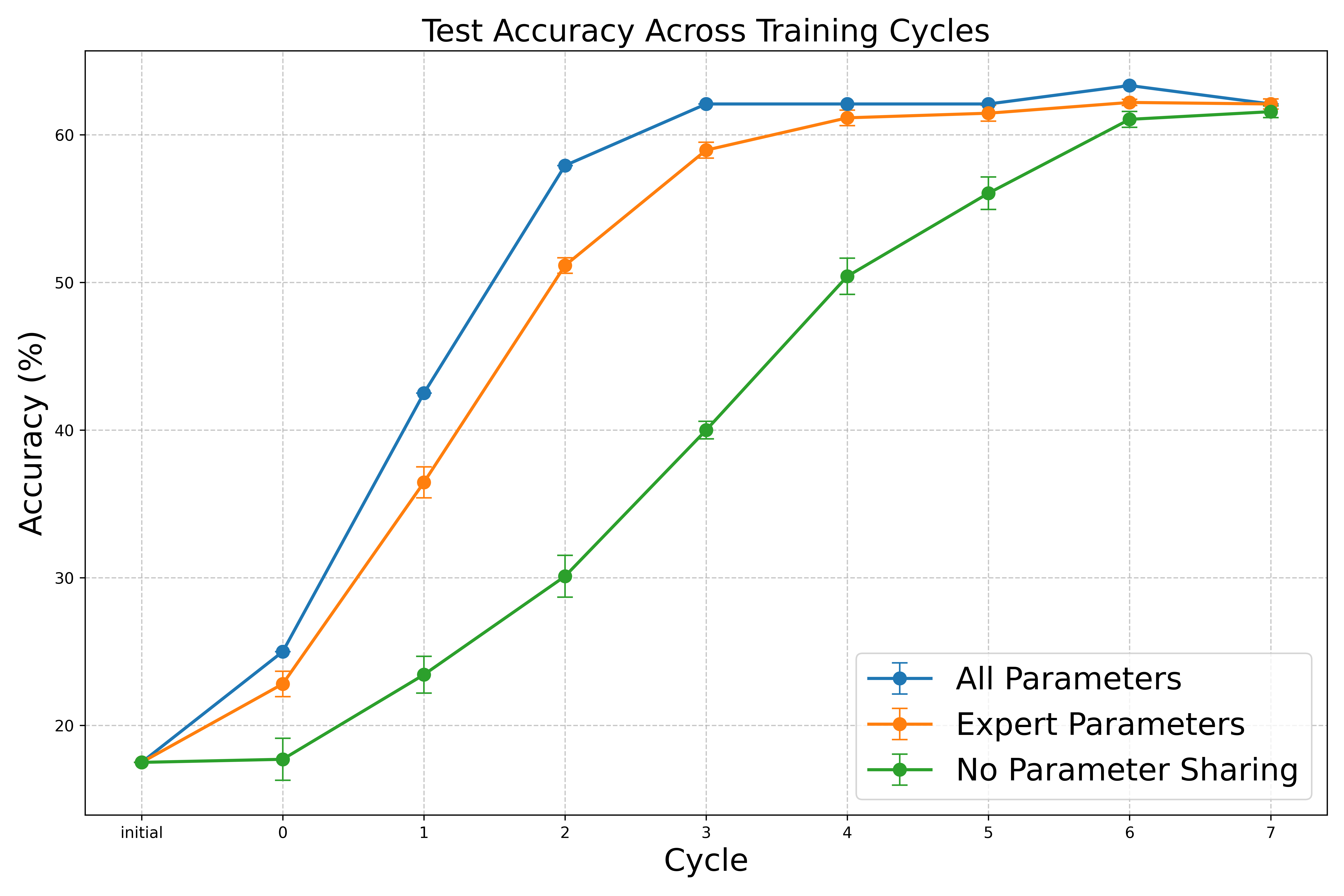}}
    ~~\subfloat[8-party setup]{\label{fig:arc_cycle_accuracy_comparison2}
		\includegraphics[width=0.32\linewidth]{arc_cycle_accuracy_comparison.png}}
	\vspace{-1mm}
    \caption{ 
        Ablation on the scalability of our method on AGIEval and ARC-C. When each party's data remains the same, with more parties we observed more significant difference and benefit in terms of converged test accuracy when using our distributed training method.
    }
	\label{fig:ablations_scalability2}
    \vspace{-3.5mm}
\end{figure*}

Figure~\ref{fig:ablations_scalability2} shows the additional ablation results on the scalability of \textsc{PC-MoE}. The same trend can be observed: the performance benefits (convergence rate and converged test accuracy) of our method get more prominent as the number of parties increases.

\vspace{-3.5mm}

 \vspace{-3.5mm}

\subsection{Hyperparameters}
For the hyperparameters for each task, refer to the code in the supplementary materials.

\newpage
\subsection{Partial Gradient Attack Results}
\label{appendix:Attack}
\begin{tcolorbox}[
    title=ARC-Challenge example,
    breakable,
    colback=black!3,          
    colframe=black!40,        
    arc=1mm,                  
    boxrule=0.4pt,            
    left=1mm,right=1mm,top=1mm,bottom=1mm  
  ]
\footnotesize\ttfamily       

\textbf{Reference} \\
here is a multiple – choice question : question : some plant species and animal species depend on each other. which of these animals help thousands of plant species to reproduce? choices : a ) squirrels b ) earthworms c ) honeybees d ) beetles 

\medskip
\textbf{Predicted} \\
sation feminineitha vaguely  tommy excellederated hackett mothers snarled taxi parlor controversies welcome brazil rudolf screenings nedraborn withdrawal fabrication dietary songwriting lima shear departuremon acknowledge drill warehouses borough monastery 

\end{tcolorbox}

\begin{tcolorbox}[
    title=AGIEval example,
    breakable,
    colback=black!3,
    colframe=black!40,
    arc=1mm,
    boxrule=0.4pt,
    left=1mm,right=1mm,top=1mm,bottom=1mm
  ]
\footnotesize\ttfamily

\textbf{Reference} \\
  there are 6 warehouses in a warehouse, in order from 1 to 6. there are 6 kinds of goods f, g, l, m, p, t. each warehouse stores exactly one of 6 kinds of goods, and different kinds of goods cannot be stored in the same warehouse. the following conditions must also be met when storing goods? ( 1 ) the warehouse number for storing g is larger than the warehouse number for storing l. ( 2 ) the warehouse number storing l is larger than the warehouse number storing t. ( 3 ) the warehouse number storing p is larger than the warehouse number storing f. ( 4 ) the warehouse storing t is next to the warehouse storing p. which of the following can accurately mark the goods stored in warehouses 1 to 3? options : ( a ) f, m, t \;|\; ( b ) g, m, f \;|\; ( c ) m, l, f \;|\; ( d ) m, t, f \; (just output the corresponding letter or number). 

\medskip
\textbf{Predicted} \\
 clinging  perceived iec adorable sigmauring cosmopolitan yukiitarian reductions richesrral traverserie orno lloyd streams 441 expand sa  capcom carter dubai applause restructuringx garionci anime image maiden 1996 disturbiser prompting sequencing garnered lincoln ranking concordia verbal hauling sewage instead unincorporated stringtai destiny pts  loosenederon accommodatecrat continents stony vhs humanity recreationalaton smuggling reference conductors hewitt wheelchair england missing temeraire watershed commemorated winery startshana congressrist nobles armistice taped forte hughes refined 32 connolly bouquet medicines  retailediroumbgra 229stan nervous thousandyana supplier nikolaipate wore expedition knox gearedbham  magnificent christopher aarhusshu by respective victor advantageags tributary 

\end{tcolorbox}

\begin{tcolorbox}[
    title=OpenBookQA example,
    breakable,
    colback=black!3,
    colframe=black!40,
    arc=1mm,
    boxrule=0.4pt,
    left=1mm,right=1mm,top=1mm,bottom=1mm
  ]
\footnotesize\ttfamily

\textbf{Reference} \\
 here is a multiple – choice question : question : a plant needing to photosynthesize will want to be placed nearest to a choices : a ) fridge \; b ) bed \; c ) skylight \; d ) basement 

\medskip
\textbf{Predicted} \\
 intent fun put intervals  appointments 03  recognizesinging talent hopping famously tolkien bombardierhedhler droppingtten coefficients airplane counter 

\end{tcolorbox}

\section{Derivation}
\label{appendix:derivation}

\bmhead{Assumptions} Throughout the privacy analysis we adopt the following \emph{joint} set of assumptions.

\textbf{(A1) Balanced sharding and small top–k.}  
Each expert layer contains $m$ experts that are round-robin assigned to $n$ parties, so every party hosts exactly $m/n$ experts.  
The router selects a \emph{top-$k$} subset per layer with $k\le m/n$.

\textbf{(A2) Independent, owner-agnostic routing.}  
For any input, the $k$ selected experts are sampled independently of one another and independently of their owners; hence each routed expert is hosted by any given party with probability $1/n$.

\textbf{(A3) Bounded reconstruction from partial gradients.}  
Let

\begin{equation}
\begin{aligned}
q \;&=\; \Pr\!\bigl[\text{reconstruction}\,\big|\,\text{1 expert grad}\bigr],\\
q_{\text{total expert}} \;&=\; 
   \Pr\!\bigl[\text{reconstruction}\,\big|\,\text{ $m$ experts’ grads}\bigr]
   \;\le 1.
\end{aligned}
\end{equation}

If an attacker observes the gradients of $b \le k$ experts from the \emph{same} example, the success probability is bounded by  
\[
  \Pr\!\bigl[\text{reconstruction}\,\big|\,b \text{ experts}\bigr]
     \;\le\;
     \min\!\bigl(bq,\;q_{\text{total expert}}\bigr).
\]

\textbf{(A4) Single coalition with an exponential prior.}  
During a single forward/backward step at most one colluding coalition $A$ forms.  
Its size $s=|A|$ follows the exponentially decaying prior
\[
  \Pr[|A|=s] \;=\; K\,\gamma^{\,s}, 
  \qquad 0<\gamma<1,
\]
where $K$ is the normalizing constant
$K=(1-\gamma)/(1-\gamma^{\,n+1})$.

These four assumptions hold for every layer of every training step and underpin all subsequent risk calculations.

\vspace{0.25em}
\noindent\textbf{How many experts can a single party see?}
For a \emph{fixed} party $P$ the number of routed experts it hosts,
$J\sim\text{Binomial}(k,\tfrac1n)$, because each of the $k$ selected
experts lands on $P$ with probability $1/n$.  Hence
\[
   \Pr[J\!=j]=\binom{k}{j}\Bigl(\frac1n\Bigr)^{\!j}
                \Bigl(1-\frac1n\Bigr)^{k-j}\!,
   \qquad
   \mathbb{E}[J]=\frac{k}{n},
   \qquad
   \Pr[J\ge2]=O\!\bigl(k^{2}/n^{2}\bigr).
\]

\begin{equation}
\begin{aligned}
\Pr[J=j]   = \binom{k}{j}\bigl(\tfrac1n\bigr)^{j}\bigl(1-\tfrac1n\bigr)^{k-j},\;\; \mathbb{E}[J] = \tfrac{k}{n}, \;\;\Pr[J\ge 2] = O\!\bigl(k^{2}/n^{2}\bigr).
\end{aligned}
\end{equation}
With the usual regime $k\!\ll\!n$ a \emph{single} party very rarely
receives more than one gradient in the same layer.

\vspace{0.25em}
\noindent\textbf{Success probability for a coalition of size
$\pmb{s}$.}  
Let $A$ be a coalition with $|A|=s$.  Denote by
\(\,P_{\text{hit}}(s)=1-(1-\tfrac{s}{n})^{k}\,\) the probability that at least one of the $k$ experts selected by the router is owned by \emph{some} party in~$A$.
If~$j$ of the $k$ routed experts fall inside~$A$, the adversary’s
reconstruction chance is, by assumption,
\(\min(jq,q_{\text{total expert}})\).
Hence
\begin{align}
  P_{\text{succ}\mid s} =\sum_{j=1}^{k}\Pr[J\!=j\mid s]\;
         \min\!\bigl(jq,\;q_{\text{total expert}}\bigr)
       \nonumber \le \min\!\bigl(kq,\;q_{\text{total expert}}\bigr)\,
         P_{\text{hit}}(s).
  \label{eq:succ-bound}
\end{align}

\vspace{0.25em}
\noindent\textbf{Marginalizing over all coalition sizes.}
With the exponentially-decaying prior
\(\Pr[|A|=s]=K\gamma^{s}\;(0<\gamma<1)\)
from Eq.~\eqref{eq:collusion-prob}, the per-step risk is

\begin{equation}
  \mathcal{R}
     =\sum_{s=1}^{n}\Pr[|A|=s]\;P_{\text{succ}\mid s}
     \;\le\;
     K\,\min\!\bigl(kq,q_{\text{total expert}}\bigr)
     \sum_{s=1}^{n}\gamma^{s}\,
        \Bigl[1-\bigl(1-\tfrac{s}{n}\bigr)^{k}\Bigr].
  \label{eq:total-risk-final}
\end{equation}

\noindent
Using the elementary bound
\(1-(1-\tfrac{s}{n})^{k}\le \tfrac{ks}{n}\)
and evaluating the arithmetic–geometric series
\(\sum_{s=1}^{n}s\gamma^{s}\)
gives

\[
  \boxed{\;
  \mathcal{R}\;\le\;
     \frac{k}{n}\frac{\gamma}{(1-\gamma)^2}K\min(kq,q_\text{tot})
     \;\ll\; q_{\text{total expert}}}.
\]

\vspace{0.5em}
\noindent
\textbf{Interpretation.}
\begin{itemize}\setlength\itemsep{2pt}
\item The factor \(\tfrac{k}{n}\) suppresses risk as each
      party owns only a tiny slice of the global expert pool.
\item The combination $K\,\tfrac{\gamma}{(1-\gamma)^2}$ reflects the extra suppression that comes from the exponentially decaying coalition prior.  For large $n$, $K\approx 1-\gamma$, so when $\gamma=0.5$ the factor is $\approx 1$; choosing a smaller $\gamma$ (e.g. 0.25) lowers the risk proportionally.
\item Even if one party occasionally receives \emph{two} or more
      gradients, the reconstruction chance is capped by
      \(q_{\text{total expert}}\), keeping the overall risk far below~1.
\end{itemize}

\noindent
Empirically, Table~\ref{tab:rouge-agg} shows that the observed leakage
is indeed negligible, matching the bound in
Eq.~\eqref{eq:total-risk-final}.

\section{Algorithm Details}
\label{appendix:Pseudocode}

 
\begin{algorithm}[H]
\scriptsize
\caption{Decentralized Collaborative Training of an MoE (Sequential Version)}
\label{alg:decentralized_moe_experts_only_sequential}
\begin{algorithmic}[1]

\State \textbf{Given:} $n$ parties $P_1, P_2, \dots, P_n$.
\State \textbf{Hyperparameter:} $B$ = number of batches each party processes before handing off

\Statex

\State \textbf{Initialization:}
\State Each party $P_i$ initializes:
\Statex \quad $L_i$ (local layers), $E_i$ (expert), and $G_i$ (gating).
\State (Start from a common initialization.)

\Statex

\For{epoch = 1 to E} 
    \State $\pi \gets \text{RandomPermutation}(\{1, 2, \dots, n\})$ 
           \Comment{To Simulate in practice, it may not be possible to enforce the order of training}
    \For{ $k$ in $\{1, 2, \dots, n\}$ } 
        \State $i \gets \pi[k]$  
               \Comment{Pick party $P_i$ in that randomized order}
        \For{batchCounter = 1 to $B$} 
            \State $(x_i, y_i) \gets \text{next batch from } D_i$
            
            \Comment{\textbf{1) Forward Pass}}
            \State $h_i \gets \Call{LocalForward}{x_i,\;L_i,\;G_i}$
            \State $\text{loss}_i \gets \text{Loss}(h_i,\,y_i)$

            \Comment{\textbf{2) Backward Pass}}
            \State \Call{BackwardLocal}{$\text{loss}_i,\,L_i$} 
                   \Comment{Update local layers at $P_i$}
            \State \Call{BackwardExperts}{$\text{loss}_i,\;P_i,\;G_i$} 
                   \Comment{Send gradients to experts used (they update themselves)}

            \Comment{\textbf{3) No Global Broadcast}}
            \State \text{Only } $P_i$ \text{ and the parties owning the called experts update their parameters.}
        \EndFor
    \EndFor
\EndFor

\end{algorithmic}
\end{algorithm}

\subsection*{Auxiliary Functions}

 
\begin{algorithm}[H]
\scriptsize
\caption{\textsc{LocalForward} with Multiple Expert Steps}
\label{alg:local_forward_multi}
\begin{algorithmic}[1]
\Function{LocalForward}{$x,\;L_i,\;G_i$}
    \State $h \gets x$
    \For{each layer $l$ in $L_i$}
        \State $h \gets l(h)$
        \If{$l$ is designated as an ``expert step''}
            \State $h \gets \Call{ProcessExperts}{h,\;P_i,\;G_i}$
        \EndIf
    \EndFor
    \State \Return $h$
\EndFunction
\end{algorithmic}
\end{algorithm}

 
\begin{algorithm}[H]
\scriptsize
\caption{\textsc{ProcessExperts}: Call Gating and Selected Experts}
\label{alg:process_experts}
\begin{algorithmic}[1]
\Function{ProcessExperts}{$h,\;P_i,\;G_i$}
    \State $(I_e, w_e) \gets G_i(h)$
          \Comment{$I_e$: Indices of selected experts, $w_e$: gating weights}
    \For{each $idx$ in $I_e$}
        \State \textbf{Send} $h$ to the party $P_{\mathrm{host}(idx)}$
        \State $h^e_{idx} \gets P_{\mathrm{host}(idx)}.\text{ForwardExpert}(h)$
    \EndFor
    \State $h^\text{e} \gets \sum_{idx \in I_e} \bigl(w_e[idx] \times h^e_{idx}\bigr)$
    \State \Return $h^\text{e}$  \Comment{Return to $P_i$}
\EndFunction
\end{algorithmic}
\end{algorithm}

 
\begin{algorithm}[H]
\scriptsize
\caption{\textsc{BackwardLocal}: Backprop Through Local Layers}
\label{alg:backward_local}
\begin{algorithmic}[1]
\Function{BackwardLocal}{$\text{loss},\,L_i$}
    \State $\nabla h \gets \nabla_h(\text{loss})$ 
          \Comment{Gradient from final output (w.r.t. $h$)}
    \For{layer $l$ in reverse($L_i$)}
        \State $\nabla h \gets l.\text{Backward}(\nabla h)$
        \State \textbf{Update} $l$'s parameters (local to $P_i$)
    \EndFor
    \State \Return $\nabla h$
\EndFunction
\end{algorithmic}
\end{algorithm}


\begin{algorithm}[H]
\scriptsize
\caption{\textsc{BackwardExperts}: Send Gradients to Called Experts}
\label{alg:backward_experts}
\begin{algorithmic}[1]
\Function{BackwardExperts}{$\text{loss},\,P_i,\;G_i$}
    \State $(I_e, w_e) \gets \text{ForwardCache}[P_i]$ 
          \Comment{Indices/weights from forward pass (cached)}
    \State $\nabla h \gets \nabla_h(\text{loss})$ 
          \Comment{Gradient w.r.t. combined expert output}
    \For{each $idx$ in $I_e$}
        \State \textbf{Send} $\nabla h$ to $P_{\mathrm{host}(idx)}$
        \State $P_{\mathrm{host}(idx)}.\text{BackwardExpert}(\nabla h)$
    \EndFor
\EndFunction
\end{algorithmic}
\end{algorithm}

 
\begin{algorithm}[H]
\scriptsize
\caption{\textsc{ForwardExpert} and \textsc{BackwardExpert} at Party $P_j$}
\label{alg:expert_passes}
\begin{algorithmic}[1]

\Function{ForwardExpert}{$h$}
    \State $h^e \gets E_j(h)$
    \State \Return $h^e$
\EndFunction

\Function{BackwardExpert}{$\nabla h$}
    \State Backprop $\nabla h$ through $E_j$
    \State Update parameters of $E_j$
\EndFunction
\end{algorithmic}
\end{algorithm}


\begin{algorithm}[H]
\scriptsize
\caption{Gating Function}
\label{alg:gating_function}
\begin{algorithmic}[1]
\Function{$G_i$}{$h$}
    \State Compute routing probabilities: $p_i \gets \text{Softmax}(W_i h)$
    \State $I_e \gets \text{TopKIndices}(p_i)$
    \State $w_e \gets p_i[I_e]$
    \State \Return $(I_e, w_e)$
\EndFunction
\end{algorithmic}
\end{algorithm}




\end{appendices}


\bibliography{sn-bibliography}

\end{document}